 \documentclass[pmlr,twocolumn,10pt]{jmlr} % W&CP article

\usepackage{booktabs}
\usepackage{siunitx}

\usepackage{placeins}

\newcommand{\cs}[1]{\texttt{\char`\\#1}}% remove this in your real article

\theorembodyfont{\upshape}
\theoremheaderfont{\scshape}
\theorempostheader{:}
\theoremsep{\newline}
\newtheorem*{note}{Note}

\jmlrvolume{225}
\jmlryear{2023}
\jmlrsubmitted{LEAVE UNSET}
\jmlrpublished{LEAVE UNSET}
\jmlrworkshop{Machine Learning for Health (ML4H) 2023} % W&CP title

\title{Robust semi-supervised segmentation\\ with timestep ensembling diffusion models}

\author{%
\Name{Margherita Rosnati} \Email{margherita.rosnati12@imperial.ac.uk}\\
\Name{Mélanie Roschewitz}\\
\Name{Ben Glocker}\\
\addr Biomedical Image Analysis Group (BioMedIA) \\ Department of Computing \\ Imperial College London
}
\begin{document}

\maketitle

\begin{abstract}
Medical image segmentation is a challenging task, made more difficult by many datasets' limited size and annotations. Denoising diffusion probabilistic models (DDPM) have recently shown promise in modelling the distribution of natural images and were successfully applied to various medical imaging tasks. 
This work focuses on semi-supervised image segmentation using diffusion models, particularly addressing domain generalisation.
Firstly, we demonstrate that smaller diffusion steps generate latent representations that are more robust for downstream tasks than larger steps. Secondly, we use this insight to propose an improved esembling scheme that leverages information-dense small steps and the regularising effect of larger steps to generate predictions.
Our model shows significantly better performance in domain-shifted settings while retaining competitive performance in-domain. Overall, this work highlights the potential of DDPMs for semi-supervised medical image segmentation and provides insights into optimising their performance under domain shift.

\end{abstract}
\begin{keywords}
Medial Image Segmentation, Semi-Supervised Learning, Generative Modelling
\end{keywords}

\section{Introduction}
Denoising diffusion probabilistic models (DDPM)~\citep{sohl2015deep,ho2020denoising} have recently emerged as a promising approach for modelling the distribution of natural images, outperforming alternative methods in terms of sample realism and diversity. More recently, DDPM have also been successfully applied to various medical imaging tasks, such as synthetic image generation~\citep{kim2022diffusion}, image reconstruction~\citep{xie2022measurement,peng2022towards},
anomaly detection~\citep{wolleb2022diffusion,pinaya2022fast}, diagnostics~\citep{aviles2022multi} and segmentation~\citep{wolleb2022diffusion}.

Image segmentation is crucial in medical imaging, where accurate and efficient methods are required to support diagnosis, treatment planning, and disease monitoring. However, medical imaging datasets are often limited in size and may lack sufficient annotations, making it challenging to train accurate segmentation models. Moreover, medical imaging data is characterised by high variability, resulting from differences in acquisition parameters, scanner types, and patient demographics. This phenomenon, also known as domain shift, poses a significant challenge to the generalisation of segmentation models applied to new datasets, leading to potential underperformance in clinical settings.

Recent research in diffusion models has shown promising results for semi-supervised learning~\citep{baranchuklabel,deja2023learning} based on the discovery that the bottleneck network, tasked to learn the backward process of removing noise from an image, also learns an expressive feature representation that can benefit other downstream analysis tasks. Several techniques have been proposed to leverage intermediate diffusion steps for improved in-domain downstream performance. However, more research is needed on the implications of these design choices regarding model generalisation. Our work focuses on the latter problem.

\begin{figure*}[!t]
% Caption and label go in the first argument and the figure contents
 % go in the second argument
\floatconts
  {fig:models_diagram}
  {\caption{Models diagram. LEDM, the SOTA in semi-supervised segmentation with diffusion models, selects a subset of timesteps and concatenates latent representations extracted from a pretrained diffusion model as features fed to an MLP. Our method (i) selects smaller and more informative timesteps, (ii) predicts through a voting mechanism over our steps selection and (ii) shares the MLP weights across timesteps, resulting in improved segmentation performance.}}
  {\includegraphics[width=\textwidth]{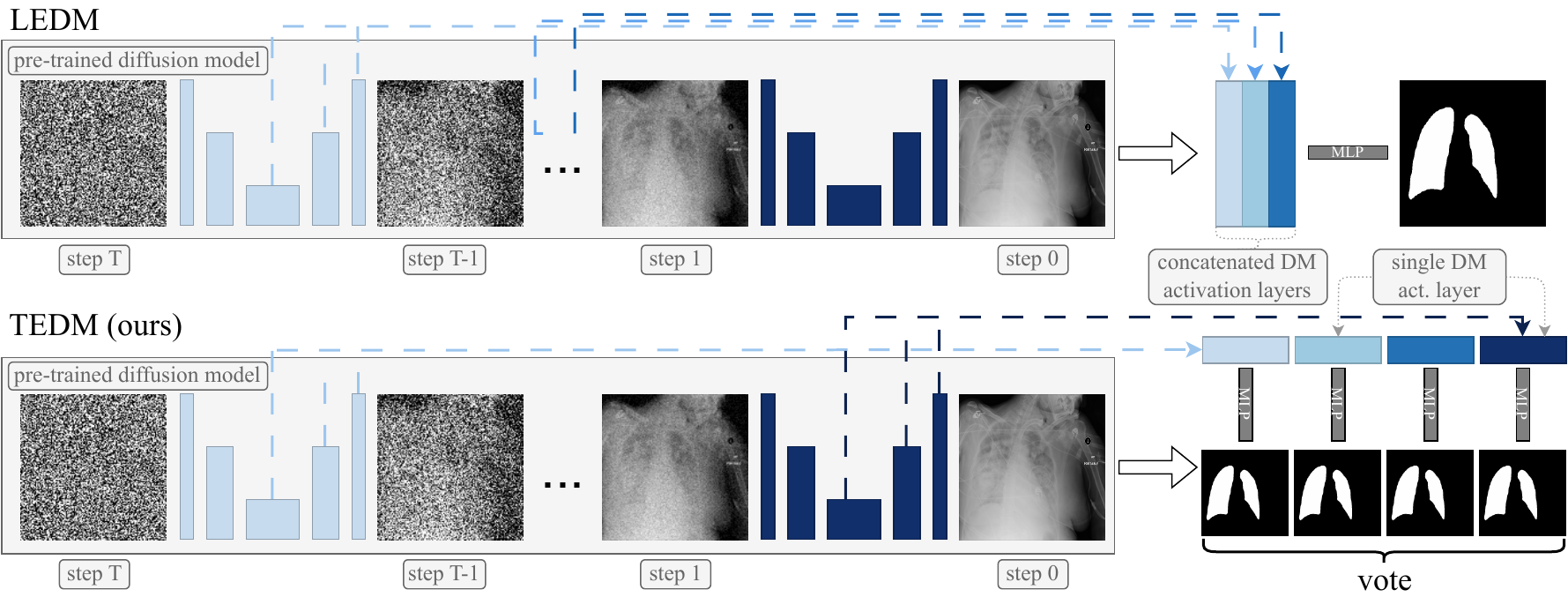}}
\end{figure*}

Specifically, we investigate how to optimally leverage diffusion steps to improve generalisation for semi-supervised image segmentation under domain shift. Based on the analysis of datasets with diverse imaging modalities and domain shifts, our findings demonstrate significant improvements over existing baselines using five different datasets. Our key findings can be summarised as follows:
%Specifically, we investigate how to optimally leverage diffusion steps to improve generalisation for semi-supervised image segmentation under domain shift, demonstrating significant improvements over existing baselines. Our key findings can be summarised as follows:
\begin{itemize}
\item Small diffusion steps are crucial for model generalisation;
\item Concatenating latent representations over steps to predict segmentation maps can hurt generalisation;
\item Instead, generalisation can be significantly improved by (i) optimising which timesteps to use at test time, (ii) ensembling predictions from individual timesteps using a shared predictor and (iii) using these individual predictions for regularisation during training.
\end{itemize}

\section{Background and related work}
\subsection{Diffusion models}
Diffusion models have garnered significant interest in the machine learning community due to their remarkable ability to model complex data distributions efficiently. 
Diffusion models utilise a series of simple and learnable transformations to diffuse noise iteratively and generate samples from the target distribution. Formally, a DDPM works as follows. Given a data distribution $p(\mathbf{x}_0)$ and forward process:
\begin{equation}
    p(\mathbf{x}_t|\mathbf{x}_{t-1}) = \mathcal{N}(\mathbf{x}_t;\sqrt{1-\beta_t} \mathbf{x}_{t-1}, \beta_t\mathbf{I}),
\end{equation}
where $\beta_t \in (0,1)$ is the variance schedule and $t\in[0,T]$ is the Markov chain time step, a DDPM aims to learn $\mathbf{\mu}_{\theta}(\mathbf{x}_t, t)$ and $\mathbf{\Sigma}_{\theta}(\mathbf{x}_t, t)$ which define the backward process:
\begin{equation}
    p(\mathbf{x}_{t-1}|\mathbf{x}_t) = \mathcal{N}(\mathbf{x}_{t-1}; \mathbf{\mu}_{\theta}(\mathbf{x}_t, t), \mathbf{\Sigma}_{\theta}(\mathbf{x}_t, t)).
\end{equation}
In order to do so, \citet{ho2020denoising} fix the variance $\mathbf{\Sigma}_{\theta}(\mathbf{x}_t, t)$, reparametrise $\mathbf{\mu}_{\theta}(\mathbf{x}_t, t)$ as a function of the noise $\mathbf{\epsilon}_{\theta}(\mathbf{x}_t, t)$
\begin{align}
   \mathbf{\mu}_{\theta}(\mathbf{x}_t, t) &= \frac{1}{\sqrt{\alpha_t}}\big(\mathbf{x}_t-\frac{1-\alpha_t}{\sqrt{1-\overline{\alpha}_t}} \mathbf{\epsilon}_{\theta}(\mathbf{x}_t, t)\big),\\\alpha_t &= 1-\beta_t, \qquad
\overline{\alpha}_t = \prod_{i=1}^t \alpha_{i}
\end{align}
and design a UNet-based~\citep{ronneberger2015u} neural network architecture
\begin{equation}
G_{\theta}: (\mathbf{x}_t,t) \rightarrow \mathbf{\epsilon}_{\theta}(\mathbf{x}_t, t)    
\end{equation}
 for learning to identify the noise. The UNet is trained through cross-entropy between the injected and predicted noise.

\subsection{Diffusion models for label-efficient image segmentation}

\citet{baranchuklabel} apply diffusion models to semi-supervised segmentation by using a diffusion model pretrained on unlabelled images, extracting latent representation from the UNet's intermediate layers and using them to train a pixel-wise classifier. More concretely, their Label Efficient Diffusion Model (LEDM) extracts the latent representations generated with a pretrained UNet diffusion model by selecting a set of steps $t\in S\subset \{0,\dots, T\}$, passing a noisy input
\begin{equation}
 \mathbf{x}_t = \sqrt{\overline{\alpha}_t}x_0 + \sqrt{1-\overline{\alpha}_t}\epsilon,\qquad\epsilon \sim \mathcal{N}(0,\mathbf{I})    
\end{equation}
through the UNet. The resulting activation maps $\mathbf{z}_t\in \mathrm{R}^{c \times h \times w}$ are then upsampled through bilinear interpolation to the input size and concatenated into a feature map $\mathbf{Z}\in \mathrm{R}^{(|S|\times c)\times H \times W}$.
Finally, each pointwise prediction is performed independently by an ensemble of lightweight multilayer perceptions 
\begin{equation}
 C_{\phi}^n: \mathbf{Z}^{i,j} \rightarrow y^{i,j};\qquad n\in\{1,..,10\}   
\end{equation}
trained with a cross-entropy loss. 
The authors concatenate the diffusion steps $S=\{50, 150, 250\}$ to form the input to these predictors.

Similarly, \citet{deja2023learning} also use the latent representations of a pretrained diffusion model for classification tasks. In particular, they propose to use classifier predictions from all intermediate timesteps to regularise the training of the diffusion model. However, at test time, they only use the last diffusion step $t=1$ to generate predictions.

\begin{figure*}[!t]
\floatconts
  {fig:results_per_timestep}
  {\caption{Performance of a logistic regression segmentation
model trained on latent features from individual diffusion steps.
}}
  {\includegraphics[width=.9\textwidth]{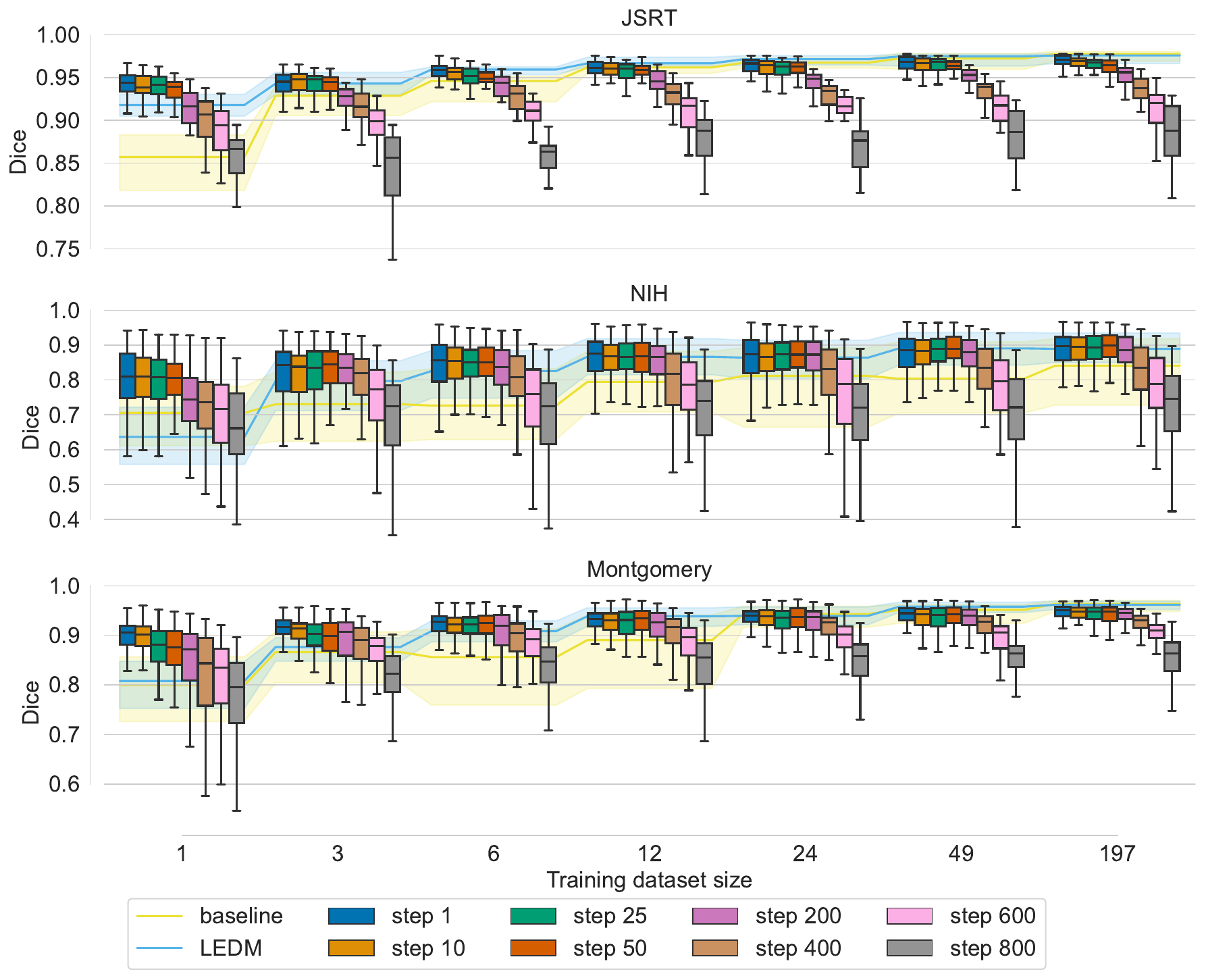}}
  
\end{figure*} 

\section{On the importance of the diffusion steps for domain generalisation}
\label{sec:diff-step-importance}

Previous findings suggest that latent representations in larger steps contain coarse information, which becomes more granular as the diffusion steps approach the target data distribution \citep{baranchuklabel,deja2023learning}. Here, we are interested in understanding how the wealth of information in each time step $s\in S$ contributes to model generalisation when the training dataset size varies.

We train a Ridge logistic regression-based pixel-wise classifier over latent representations extracted from specific timesteps $t=\{\numlist{1;10;25;50;200;400;600;800}\}$ to isolate the predictive power of each timestep. We compare these timestep-wise predictions to LEDM and a fully supervised baseline using the same UNet backbone as the DDPM backbone.

We evaluate our work on the task of chest X-ray lung segmentation. Chest X-rays are among the most frequent radiological examinations in clinical practice, and automatically extracted features from anatomical regions such as the lungs can aid clinical decision-making. Moreover, the availability of several public datasets of chest X-ray images allows us to investigate the methods’ generalisation ability in the presence of changes in dataset characteristics.

Following previous work in semi-supervised medical image segmentation~\citep{rosnati2022analysing}, we use the ChestX-ray8~\citep{wang2017chestx} (n=108k) as the unlabelled dataset to train the DDPM backbone over $T=\num{1000}$ steps and a subset of the JSRT~\citep{van2006segmentation} (n=247) labelled dataset for training (n=197) and validating (n=25) our method. The dataset splits, architecture, and code are available in our code repository\footnote{Demo: \url{https://huggingface.co/spaces/anonymous2023-21/TEDM-demo}}.

We reserve the remaining JSRT samples (n=25) along with the NIH~\citep{tang2019xlsor} (n=95), and Montgomery~\citep{jaeger2014two} (n=138) labelled datasets for final testing. Notably, the NIH dataset is an annotated subset of the ChestX-ray8 dataset. This setup allows us to test the models on data that is (i) in-domain for the classifier (JSRT), (ii) out-of-domain for the classifier but in-domain for the DDPM (ChesX-ray8/NIH) and (iii) out-of-domain for both (Montgomery).

 \figureref{fig:results_per_timestep} shows the Dice coefficients\footnote{Dice $=2\frac{|A\cup B|}{|A|+|B|}$} from the step-wise experiment when training our segmentation model, the baseline and LEDM on $n=\{$\numlist{197;49;24;12;6;3;1}$\}$ JSRT labelled datapoints, corresponding to \{\numlist{100;50;25;12;6;3;2;1}\} \% of the training dataset. Surprisingly, LEDM does not significantly\footnote{Significance is calculated through a Wilcoxon paired test at level 0.05.} outperform the baseline in the one-shot setting for domain-shifted datasets (NIH, Montgomery). This indicates that LEDM may not fully utilise the latent representation information. 
Secondly, we find that the predictor trained on a single step $t=1$ statistically outperforms both LEDM and the baseline for small training sizes (1, 3, 6 in NIH and Montgomery and for one datapoint in JSRT). In addition, this predictor remains competitive with both the baseline and LEDM across all other training dataset sizes.

%The experiment highlights that latent representations from smaller steps with more fine-grained details are more powerful predictors than coarser latent representations, particularly for domain generalisation. In particular, the choice of steps of LEDM, \{\numlist{50;125;250}\} is suboptimal for segmentation as single-step approaches with smaller steps outperform LEDM  on out-of-distribution datasets.
The experiment highlights that latent representations obtained from smaller steps are more powerful predictors than those obtained from larger steps, particularly for domain generalisation. In particular, the LEDM steps {50, 125 and 250} are not the optimal choice for segmentation as single-step approaches with smaller steps perform better on out-of-distribution datasets. 
In the next section, we investigate whether ensembling different steps can still outperform single-step approaches given the right choice of steps. We investigate several ways of ensembling these steps and their impact on model generalisation.

\begin{table*}[!t]
\floatconts
{tab:res}
{\caption{Models performance w.r.t.\ ground truth segmentations. Reported as mean $\pm$ standard deviation over the dataset. Global CL, Global \& Local CL and LEDM are a reproduction of \citet{chen2020simple}, \citet{chaitanya2020contrastive} and \citet{baranchuklabel} respectively. All statistically comparably best performing models are highlighted in bold. Significance is calculated through a Wilcoxon paired test at level 0.05.}}
{
\setlength{\tabcolsep}{6pt}
\begin{tabular}{@{}lccccc@{}}
\toprule
Training size            & 1 (1\%) & 3 (2\%) & 6 (3\%) & 12 (6\%) & 197 (100\%)\\

\midrule
&\multicolumn{5}{c}{JSRT (in-domain for classifier)}                                                                                                                                                                                                                                     \\ 
\cmidrule{2-6}

Sup. Baseline&	84.4 $\pm$ 5.4&	91.7 $\pm$ 3.7&	93.3 $\pm$ 2.9&	95.3 $\pm$ 2.3&	97.3 $\pm$ 1.2\\
Global CL&88.8 $\pm$ 5.9&92.7 $\pm$ 1.8&93.6 $\pm$ 1.6&95.3 $\pm$ 1.1&97.1 $\pm$ 1.4\\
Global \& Local CL&89.8 $\pm$ 5.2&93.1 $\pm$ 1.7&92.9 $\pm$ 1.9&94.8 $\pm$ 1.49&97.2 $\pm$ 1.2\\
%Step 1 (linear)&	\textbf{94.0 $\pm$ 1.9}&	94.3 $\pm$ 1.6&	95.6 $\pm$ 1.3&	95.9 $\pm$ 1.3&	96.7 $\pm$ 1.2\\
%Step 1 (MLP)&	91.1 $\pm$ 5.0&	\textbf{94.5 $\pm$ 2.1}&	\textbf{96.0 $\pm$ 1.4}&	\textbf{96.8 $\pm$ 1.1}&	\textbf{97.4 $\pm$ 1.3}\\
LEDM&	90.8 $\pm$ 3.5&	94.1 $\pm$ 1.6&	95.5 $\pm$ 1.4&	96.4 $\pm$ 1.4&	97.0 $\pm$ 1.3\\
LEDMe&	\textbf{93.7 $\pm$ 2.6}&	\textbf{95.5 $\pm$ 1.5}&	\textbf{96.7 $\pm$ 1.5}&	\textbf{97.0 $\pm$ 1.1}&	\textbf{97.6 $\pm$ 1.2}\\
TEDM (ours)&	\textbf{93.1 $\pm$ 3.4}&	94.8 $\pm$ 1.4 &	95.8 $\pm$ 1.2&	96.6 $\pm$ 1.1&	97.3 $\pm$ 1.2\\

\midrule
&\multicolumn{5}{c}{NIH (in-domain for DDPM, OOD for classifier)}                                                                                                                                                                                                                                    \\ 
\cmidrule{2-6}

Sup. Baseline&	68.5 $\pm$ 12.8&	71.2 $\pm$ 15.1&	71.4 $\pm$ 15.9&	77.8 $\pm$ 14.0&	81.5 $\pm$ 12.7\\
Global CL &70.7 $\pm$ 14.6&80.3 $\pm$ 12.2&77.1 $\pm$ 16.4&84.6 $\pm$ 10.8&86.9 $\pm$ 10.8\\
Global \& Local CL &71.1 $\pm$ 16.2&79.6 $\pm$ 12.7&81.1 $\pm$ 14.0&82.2 $\pm$ 13.6&87.4 $\pm$ 10.8\\
%Step 1 (linear)&	\textbf{80.4 $\pm$ 8.6}&	81.7 $\pm$ 8.2&	84.4 $\pm$ 7.6&	86.3 $\pm$ 6.7&	88.7 $\pm$ 5.4\\
%Step 1 (MLP)&	70.4 $\pm$ 10.9&	78.9 $\pm$ 9.4&	84.2 $\pm$ 8.3&	87.5 $\pm$ 6.5&	91.9 $\pm$ 3.3\\
LEDM&	63.3 $\pm$ 12.2&	78.0 $\pm$ 10.1&	81.2 $\pm$ 9.3&	85.9 $\pm$ 7.4&	88.9 $\pm$ 5.9\\
LEDMe&	70.3 $\pm$ 11.4&	78.3 $\pm$ 9.8&	83.0 $\pm$ 8.6&	84.4 $\pm$ 8.1&	90.1 $\pm$ 5.3\\
TEDM (ours)&	\textbf{80.3 $\pm$ 9.0}&	\textbf{86.4 $\pm$ 6.2}&	\textbf{89.2 $\pm$ 5.5}&	\textbf{91.3 $\pm$ 4.1}&	\textbf{92.9 $\pm$ 3.2}\\
\midrule
&\multicolumn{5}{c}{Montgomery (OOD for DDPM and classifier)}                                                                                                                                                                                                                                    \\ 
\cmidrule{2-6}
Sup. Baseline&	77.1 $\pm$ 12.0&	83.0 $\pm$ 12.2&	80.9 $\pm$ 14.7&	83.8 $\pm$ 14.9&	94.1 $\pm$ 6.6\\
Global CL &76.1 $\pm$ 15.0	&87.6 $\pm$ 9.7	&88.8 $\pm$ 11.4	&90.4 $\pm$ 10.4	&92.9 $\pm$ 10.8	\\
Global \& Local CL &77.4 $\pm$ 17.4	&88.7 $\pm$ 9.14	&89.9 $\pm$ 8.2	&90.1 $\pm$ 10.9	&92.5 $\pm$ 11.2	\\
%Step 1 (linear)&	89.5 $\pm$ 3.9&	90.8 $\pm$ 4.1&	91.8 $\pm$ 3.9&	92.3 $\pm$ 3.9&	93.9 $\pm$ 4.0\\
%Step 1 (MLP)&	85.9 $\pm$ 4.0&	89.3 $\pm$ 4.2&	92.2 $\pm$ 4.2&	93.9 $\pm$ 3.9&	94.9 $\pm$ 5.3\\
LEDM&	79.3 $\pm$ 8.1&	85.9 $\pm$ 7.4&	89.4 $\pm$ 6.7&	92.3 $\pm$ 7.2&	94.4 $\pm$ 7.2\\
LEDMe&	80.7 $\pm$ 6.6&	86.3 $\pm$ 6.5&	89.5 $\pm$ 5.9&	91.2 $\pm$ 5.6&	\textbf{95.3 $\pm$ 4.0}\\
TEDM (ours)&	\textbf{90.5 $\pm$ 5.3}&	\textbf{91.4 $\pm$ 6.1}&	\textbf{93.3 $\pm$ 6.0}&	\textbf{94.6 $\pm$ 6.0}&	95.1 $\pm$ 6.9\\
\bottomrule
\end{tabular}
}
\end{table*}

\section{Timestep ensembling diffusion models}
\label{sec:improving LEDM}

In this section, we show that the generalisation of diffusion-based segmentation models in the low data regime can be significantly improved by judiciously combining adequate timesteps both at prediction and training time. 

We hypothesise that the lack of generalisation of LEDM observed in the previous section can be mitigated with more model regularisation and reducing the number of parameters that need to be learned. Indeed, the current approach of concatenating features from numerous timesteps to feed into the pixel-wise MLP predictor results in an excessively high-dimensional input, which leads to a complex predictor. To address this concern, we propose using a shared MLP trained to generate a prediction map from each latent representation of the steps considered.

\begin{figure*}[!t]
\floatconts
    {fig:visualsation}
    {\caption{Segmentation examples. Col. 1 and 2 are the image and ground truth segmentation. Subsequent columns correspond to models trained with $n$ training datapoints (see title). Row 1 corresponds to the baseline outcomes, and row 2, 3 and 4 to LEDM, LEDMe and TEDM (our method) respectively.}}

  {%
    \subfigure[JSRT (LHS) and NIH (RHS), where NIH is OOD for the classifier. ]{\label{fig:chest x-ray vis}%
      \includegraphics[width=\textwidth]{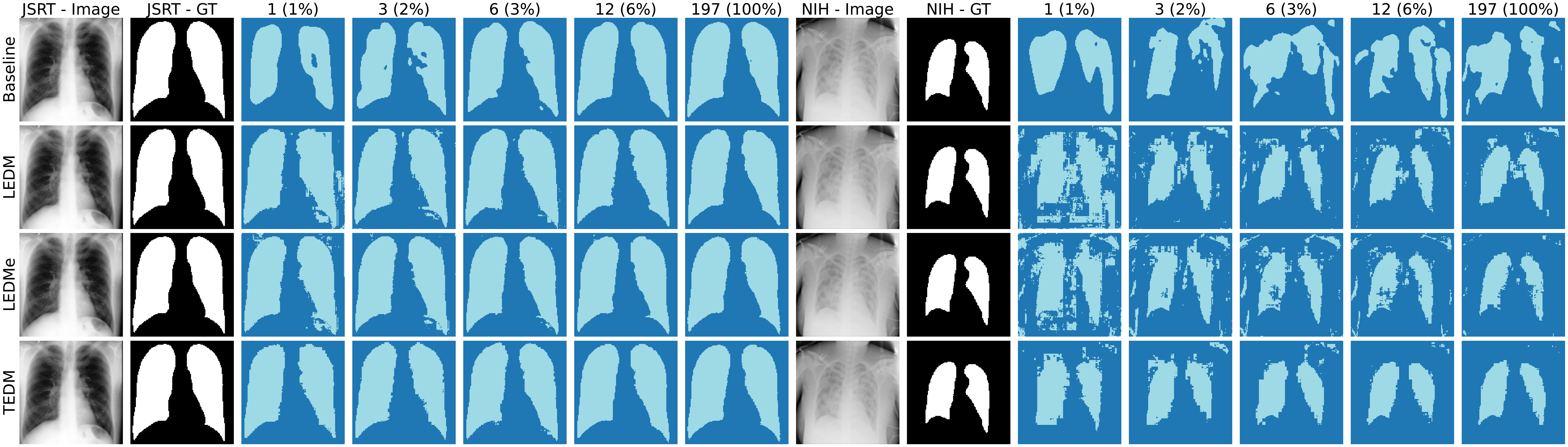}}%
    
    \subfigure[UK Biobank]{\label{fig:brats vis}%
     \includegraphics[width=.5\textwidth]{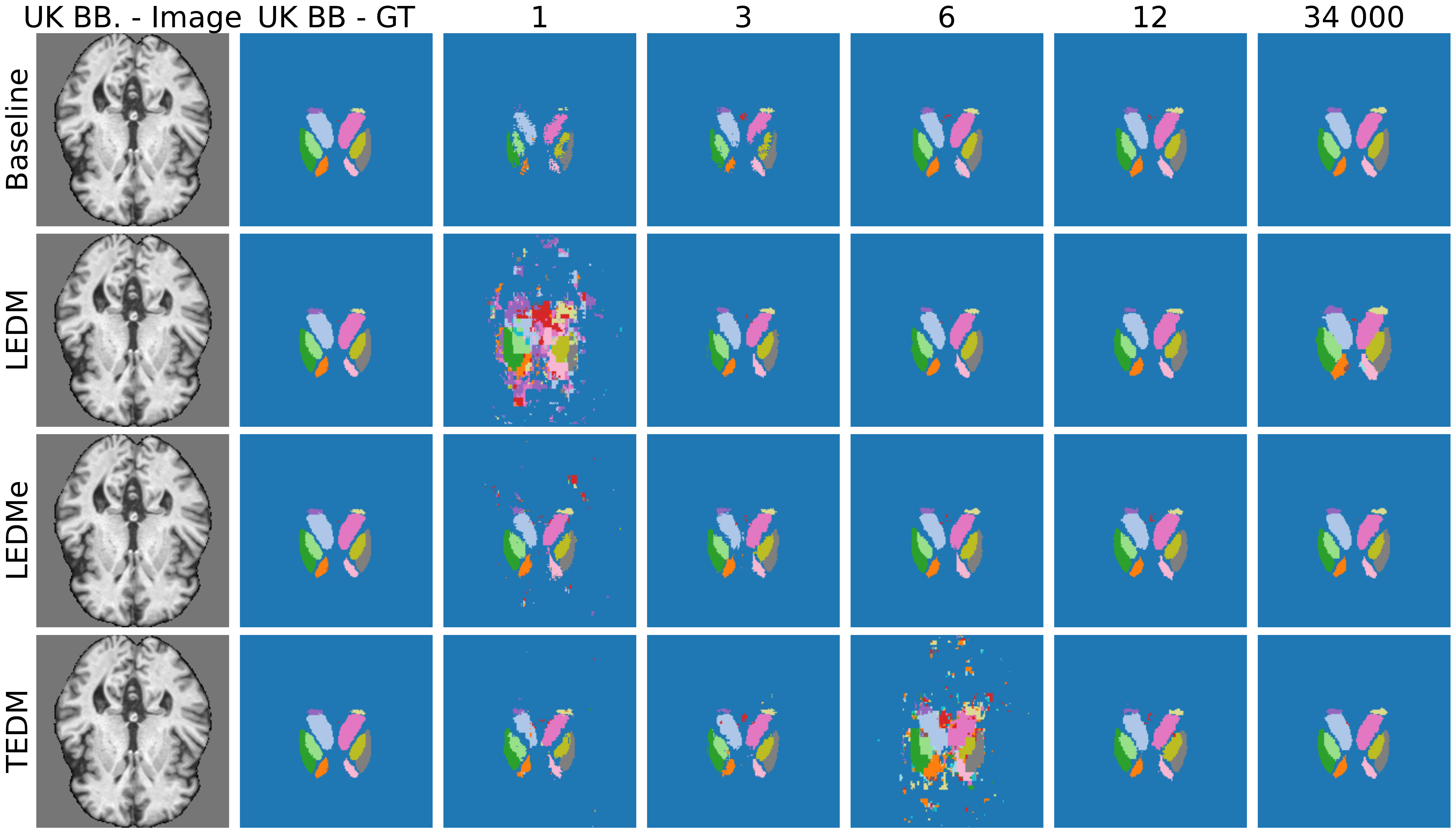}}
      \subfigure[BraTS]{\label{fig:bb vis}%
     \includegraphics[width=.5\textwidth]{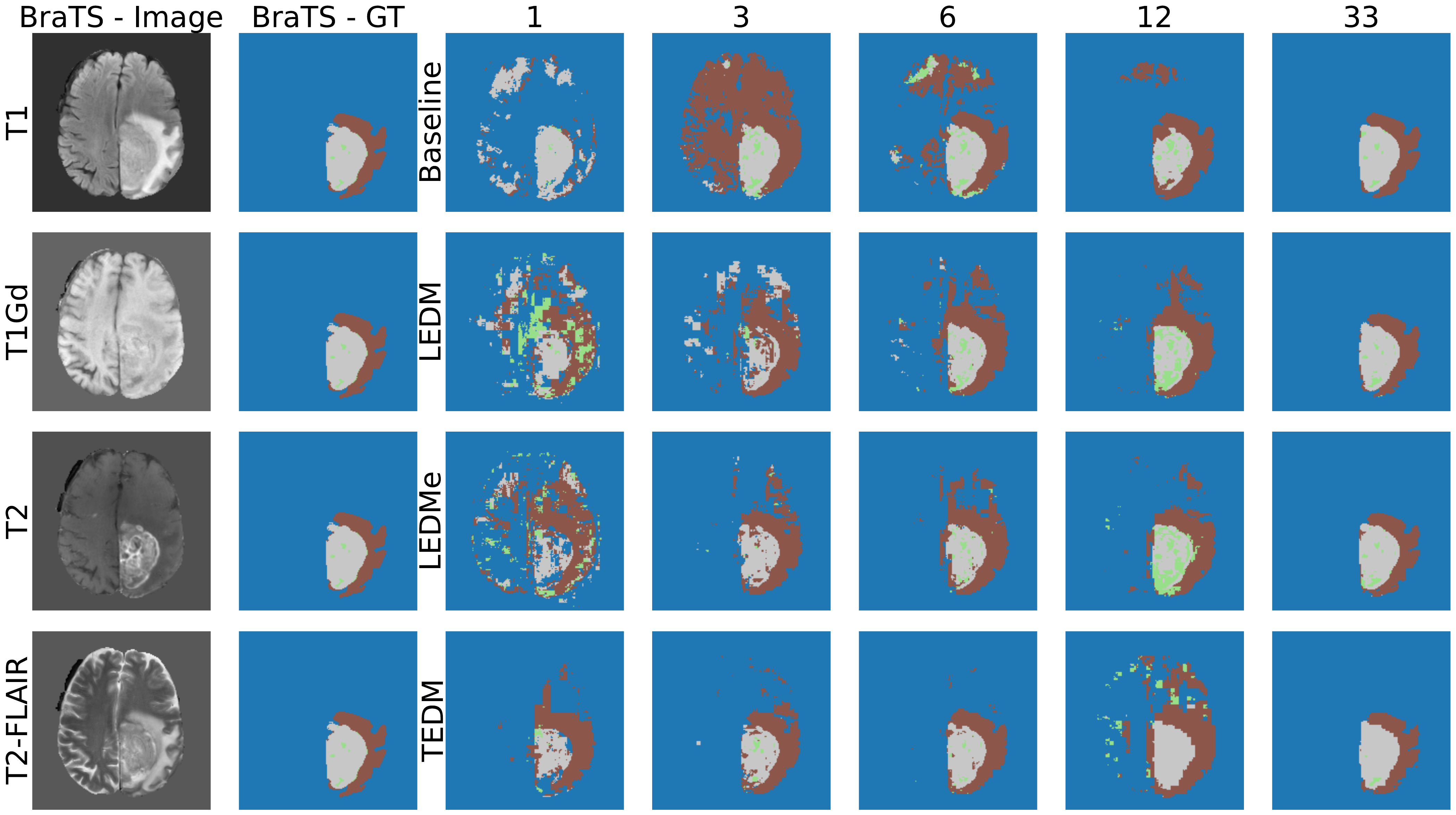}}
  }
    
\end{figure*}

%We hypothesise that the lack of generalisation of LEDM observed in the previous section can be mitigated with more model regularisation and a predictor with fewer parameters. Indeed, concatenating features from numerous timesteps to feed into the pixel-wise MLP predictor leads to an excessively high-dimensional input, resulting in a complex predictor. To address this concern, we propose using a shared MLP that generates prediction maps from the latent representation of every step considered. 
We define our loss function as follows:
\begin{equation}
   \phi = \text{argmin } \mathbb{E}_{\mathcal{D}}\mathbb{E}_{i,j}\mathbb{E}_{s\in S}\text{ CE }(C_{\phi}(\tilde{\mathbf{z}}_s^{i,j}), y^{i,j}),
\end{equation}
where $i,j$ is the pixel indexing, $y^{i,j}$ is the ground truth class of pixel $i,j$, $\tilde{\mathbf{z}}_s$ is the upsampled latent representation $\mathbf{z}_s$ of the diffusion model at step $s$, $S$ is the set of diffusion steps used, $C_{\phi}$ is the pixel-wise MLP predictor, CE stands for cross entropy and $\mathcal{D}$ is the training set.
At test time, we use a voting mechanism to ensemble the various prediction maps to obtain a final segmentation map. We call this technique ``timestep ensembling" and show that it yields superior performance.
\begin{equation}
\hat{y}_{i,j} = \frac{1}{|S|}\sum_{t\in S} C_{\phi}(\tilde{\mathbf{z}}_s^{i,j})
\end{equation}
Moreover, we leverage the insights from the previous section and combine predictions from the diffusion steps $S=$~\{\numlist{1;10;25;50;200;400;600;800}\}.
This approach allows us to benefit from the small steps information content and larger step regularisation effect, unlike LEDM, which only used timesteps \{\numlist{50;125;250}.
%and reduce training times compared to~\cite{deja2023learning}.  
To better understand the distinctions between our model and LEDM, please refer to~\figureref{fig:models_diagram}.
A discussion on computational complexity can be found in Appendix Section~\ref{Appendix:comp-cost}.

\begin{table*}[!t]
\floatconts
{tab:ablation}
{\caption{Ablation study on test-time ensembling over timesteps. Each `Step \textit{i}' experiment only uses predictions from timestep \textit{i} at test time. All statistically comparably best performing models are highlighted in bold. Significance is calculated through a Wilcoxon paired test at level 0.05.}}
{\setlength{\tabcolsep}{6pt}
\begin{tabular}{@{}lccccc@{}}
\toprule
Training size & 1 (1\%) & 3 (2\%) & 6 (3\%) & 12 (6\%) & 197 (100\%)\\

\midrule
&\multicolumn{5}{c}{JSRT (in-domain for classifier)}                                                                                                                                                                                                                                    \\ 
\cmidrule{2-6}

Step 1&	91.1 $\pm$ 5.0&	\textbf{94.5 $\pm$ 2.1}&	\textbf{96.0 $\pm$ 1.4}&	\textbf{96.8 $\pm$ 1.1}&	\textbf{97.4 $\pm$ 1.3}\\
Step 10&	91.6 $\pm$ 4.6&	\textbf{94.6 $\pm$ 1.8}&	\textbf{96.0 $\pm$ 1.3}&	\textbf{96.9 $\pm$ 1.0}&	\textbf{97.4 $\pm$ 1.2}\\
Step 25&	91.7 $\pm$ 4.2&	\textbf{94.5 $\pm$ 1.6}&	95.8 $\pm$ 1.2&	96.8 $\pm$ 1.0&	97.3 $\pm$ 1.2\\
TEDM&	\textbf{93.1 $\pm$ 3.4}&	\textbf{94.8 $\pm$ 1.4}&	95.8 $\pm$ 1.2&	96.6 $\pm$ 1.1&	\textbf{97.3 $\pm$ 1.2}\\
\midrule
&\multicolumn{5}{c}{NIH (in-domain for DDPM, OOD for classifier)}                                                                                                                                                                                                                                  \\ 
\cmidrule{2-6}

Step 1&	70.4 $\pm$ 10.9&	78.9 $\pm$ 9.4&	84.2 $\pm$ 8.3&	87.5 $\pm$ 6.5&	91.9 $\pm$ 3.3\\
Step 10&	73.2 $\pm$ 10.3&	81.1 $\pm$ 8.3&	85.8 $\pm$ 7.3&	88.8 $\pm$ 5.6&	91.8 $\pm$ 3.3\\
Step 25&	75.1 $\pm$ 9.8&	82.6 $\pm$ 7.7&	86.5 $\pm$ 6.7&	89.4 $\pm$ 5.2&	91.9 $\pm$ 3.3\\
TEDM&	\textbf{80.3 $\pm$ 9.0}&	\textbf{86.4 $\pm$ 6.2}&	\textbf{89.2 $\pm$ 5.5}&	\textbf{91.3 $\pm$ 4.1}&	\textbf{92.9 $\pm$ 3.2}\\
\midrule
&\multicolumn{5}{c}{Montgomery (OOD for DDPM and classifier)}                                                                                                                                                                                                                                     \\ 
\cmidrule{2-6}

Step 1&	85.9 $\pm$ 4.0&	89.3 $\pm$ 4.2&	92.2 $\pm$ 4.2&	93.9 $\pm$ 3.9&	94.9 $\pm$ 5.3\\
Step 10&	87.1 $\pm$ 4.5&	89.3 $\pm$ 4.8&	92.1 $\pm$ 5.2&	94.1 $\pm$ 5.0&	94.8 $\pm$ 6.5\\
Step 25&	87.4 $\pm$ 5.3&	89.1 $\pm$ 5.5&	91.7 $\pm$ 6.2&	93.7 $\pm$ 6.3&	94.6 $\pm$ 7.0\\
TEDM&	\textbf{90.5 $\pm$ 5.3}&	\textbf{91.4 $\pm$ 6.1}&	\textbf{93.3 $\pm$ 6.0}&	\textbf{94.6 $\pm$ 6.0}&	\textbf{95.1 $\pm$ 6.9}\\
\bottomrule
\end{tabular}}
\end{table*}
\section{Experiments}
We conduct experiments on various percentages of the JSRT training dataset, 12\%, 6\%, 3\%, 2\%, and 1\%, to fully explore the potential of our semi-supervised method. In addition, we train on 100\% of the training set for completeness. 
To evaluate the performance of our timestep ensembling diffusion model (TEDM), we compare it with the fully supervised baseline (described in Section~\ref{sec:diff-step-importance}) and LEDM. LEDM and TEDM have the same MLP classifier architecture. In addition, we compare TEDM to two other semi-supervised methods that use contrastive learning (CL): the ‘Global CL’~\citep{chen2020simple} and the ‘Local and Global CL’~\citep{chaitanya2020contrastive}. Both these methods are trained with the same backbone architecture as the baseline and the DDPM.

In order to investigate the effect of each component in our TEDM model, we carry out several ablations. Firstly, we compare the original LEDM model with another instance of LEDM, trained with our diffusion steps, which we refer to as LEDMe. This allows us to ablate the effect of our diffusion steps choice. Secondly, we test the voting mechanism by reporting model performance when only steps 1, 10 or 25 are used at test time. We use the same evaluation procedure as in Section~\ref{sec:diff-step-importance}. 

Finally, to test the TEDM method's generalizability, we apply it to two additional datasets: the UK Biobank dataset and the BraTS dataset~\citep{menze2014multimodal,bakas2017advancing,bakas2018identifying}. In the UK Biobank dataset, we segment brain structures in 2D slices of brain MRI T1 images. This dataset is particularly challenging due to the low intensity variation between structures and background. The BraTS dataset comprises brain MRI (T1, T1Gd, T2 and T2-FLAIR) of patients with brain tumours, which we decompose into 2D slices and segment. This dataset is even more difficult as it entails segmenting items of varied shapes and locations. Further details on the experimental process for these two datasets are available in Appendix~\ref{apd:first}.

% added to table caption
%The outcome of the Wilcoxon test is incorporated by highlighting in bold all results statistically equivalent to the best result for each experiment.

\begin{table*}[!t]
\floatconts
{tab:biobank_brats}
{\caption{Dice scores on the UK Biobank and BraTS datasets. For both datasets, the model was trained on 2D slices, the results are reported on the 3D images. The training size refers to the number of patients in the labelled training set. The number of 2D slices is roughly 100x larger. All statistically comparably best performing models are highlighted in bold. Significance is calculated through a Wilcoxon paired test at level 0.05 with Bonferroni correction to account for multiple classes per patient.
}}
{
\setlength{\tabcolsep}{6pt}
\begin{tabular}{@{}lccccc@{}}
&\multicolumn{5}{c}{UK Biobank ($n^{unlabelled}_{train} =$ \num{34000}, $n_{test} = 500$)}\\ 
\toprule
Training size            & 1 & 3 & 6 & 12 & \num{34000}\\
\midrule
%\cmidrule{2-6}
Sup. Baseline 	& 54.6 $\pm$ 18.6 & 76.8 $\pm$ 12.3 & \textbf{83.1 $\pm$ 8.5} & \textbf{85.1 $\pm$ 7.6} & \textbf{89.6 $\pm$ 5.2} \\
%\cmidrule{2-6}
Global CL 	& 42.7 $\pm$ 20.4 & 77.3 $\pm$ 11.0 & 82.0 $\pm$ 8.7 & 85.2 $\pm$ 7.4 & 88.7 $\pm$ 5.6 \\
Global \& Local CL 	& 44.3 $\pm$ 20.3 & 74.0 $\pm$ 11.8 & 80.6 $\pm$ 9.4 & 82.0 $\pm$ 8.9 & 87.4 $\pm$ 6.8 \\
LEDM 	& 60.8 $\pm$ 17.1 & \textbf{81.3 $\pm$ 7.9} & 82.3 $\pm$ 8.9 & 83.0 $\pm$ 9.2 & 87.7 $\pm$ 5.8 \\
LEDMe 	& 54.7 $\pm$ 17.8 & 79.4 $\pm$ 10.8 & 82.5 $\pm$ 9.1 & 83.8 $\pm$ 8.6 & 86.6 $\pm$ 7.0 \\
%\cmidrule{2-6}
TEDM (ours) 	& \textbf{71.0 $\pm$ 14.8} & \textbf{81.0 $\pm$      9.0} & \textbf{82.8 $\pm$ 8.8} & 83.2 $\pm$ 9.3 & 85.1 $\pm$ 7.4 \\
\bottomrule
&\multicolumn{5}{c}{ } \\ 
&\multicolumn{5}{c}{BraTS ($n^{unlabelled}_{train} = 268$, $n_{test} = 33$)} \\ 
\toprule
Training size            & 1 & 3 & 6 & 12 & \num{33}\\
\midrule%\cmidrule{2-6}
Sup. Baseline 	& 12.5 $\pm$ 18.9 & \textbf{30.9 $\pm$ 31.2} & \textbf{40.7 $\pm$ 33.1} & \textbf{47.1 $\pm$ 33.8} & \textbf{69.5 $\pm$ 25.7} \\
%\cmidrule{2-6}
Global CL 	& 4.7 $\pm$ 13.6 & 25.5 $\pm$ 29.4 & 32.3 $\pm$ 32.1 & 40.5 $\pm$ 32.0 & 56.9 $\pm$ 28.6 \\
Global \& Local CL 	& 11.7 $\pm$ 19.1 & 27.3 $\pm$ 30.5 & 34.1 $\pm$ 31.5 & 38.3 $\pm$ 32.2 & 55.4 $\pm$ 30.0 \\
LEDM 	& 24.0 $\pm$ 22.9 & 31.0 $\pm$ 31.4 & 40.8 $\pm$ 31.9 & \textbf{48.0 $\pm$ 31.2} & 62.6 $\pm$ 26.7 \\
LEDMe 	& 21.2 $\pm$ 22.7 & 33.1 $\pm$ 31.4 & \textbf{42.8 $\pm$ 32.7} & \textbf{49.5 $\pm$ 31.7} & 63.2 $\pm$ 27.6 \\
%\cmidrule{2-6}
TEDM (ours) 	& \textbf{27.3 $\pm$ 26.1} & \textbf{35.6 $\pm$ 31.7} & \textbf{41.9 $\pm$ 32.3} & \textbf{47.5 $\pm$ 31.7} & 59.8 $\pm$ 29.0 \\
\bottomrule

\end{tabular}}
\end{table*}

\section{Results}
The performance results on chest X-rays and brain MRI are shown quantitatively in Tables~\ref{tab:res} and ~\ref{tab:biobank_brats}, and qualitatively in Figure~\ref{fig:visualsation}. The ablation results are shown in Table~\ref{tab:ablation}. Further results can be found in Appendix~\ref{apd:second}. For all tables, the best-performing model and all statistically equivalent models are highlighted by reporting their results in bold.

\subsection*{Using small step sizes improves performance both in- and out-of-domain.}
 In Table~\ref{tab:res}, we observe that in all cases, selecting small diffusion steps generates the best-performing models: LEDMe outperforms LEDM statistically significantly for all experiments but two (Montgomery $n=12$ and NIH $n=12$). In addition, LEDMe outperforms LEDM for the UK Biobank and BraTS datasets for training sizes larger than 3 and 1, respectively (see Table~\ref{tab:biobank_brats}). 

\subsection*{Concatenating latent representations hurts generalisability in the low data regime.}
%Both Step 1 and O
TEDM outperforms LEDMe (and LEDM) for the NIH and Montgomery datasets, except for n=197. We deduce that the concatenation method exploited in LEDM leads to poor generalisation on domains outside the labelled training set.
In addition, TEDM performs statistically comparably to LEDM for JSRT, indicating that its generalisation properties come with little to no in-domain performance cost.

\subsection*{Test-time ensembling over timesteps improves generalisation over single-step predictions.}
%For both NIH and Montgomery, our method outperforms all other methods, except for Montgomery with the full dataset, where LEDMe outperforms it.% (and for NIH-\{1\}, where it performs similarly to Step 1).
Table~\ref{tab:ablation} shows that using a voting mechanism for prediction (used in TEDM) is more effective than using the smallest step (TEDM outperforms the competing models in OOD cases), implying that different steps produce latent representations focusing on slightly different aspects of the image.

\subsection*{TEDM performs robustly for increasingly challenging segmentation tasks.}
Table~\ref{tab:biobank_brats} shows that TEDM is statistically superior or equal to its competitors for all cases with less than 12 datapoints, showing that our method remains competitive in more challenging in-domain low labelled data scenarios.

\subsection*{Fully supervised baselines are competitive for in-domain harder segmentation tasks.}
Our method TEDM  showcases excellent performance on very small dataset sizes (\numlist{1;2;3;6} in Table~\ref{tab:biobank_brats}). However, for larger datasets (6 patients or more), a well-designed baseline model proves to be more effective than any of the semi-supervised models.
This result suggests that although semi-supervised methods with self-supervised pretraining may have their limitations in providing task-specific performance for larger datasets, they present great potential for improving results on small datasets.

%The results indicate that latent representations across different steps share semantics and act as a model regulariser, leading to better generalisation than competing methods.
%Although our method TEDM performs better for very small dataset sizes (\numlist{1;2;3;6} in Table~\ref{tab:biobank_brats}), a well fitted baseline is competitive for larger datasets (6 patients or more). This result suggests that self-supervised pretraining on the scale of a single dataset\footnote{as opposed to self supervised pretraining seen in foundation models} has limited scope when the supervision is not tailored further to the downstream task.

\section{Conclusions}

This study investigated the impact of different diffusion steps on the performance and generalisation of semi-supervised segmentation models. Our comprehensive experiments across multiple datasets revealed that small diffusion steps are crucial for domain generalisation, requiring only a few training samples to become powerful pixel-wise predictors. 
Furthermore, we found that ensembling segmentation maps over timesteps significantly improves model generalisation in the low data regime while offering competitive performance in-domain. 
Conversely, concatenating latent representations can hurt the generalisation of the pixel-wise classifier. 
These findings were demonstrated by the superior performance of our proposed Timestep Ensembling Diffusion Model on chest X-ray lung segmentation and more challenging tasks such as brain structure and tumour segmentation. 
Our results indicate that latent representations across different steps share semantics and act as a model regulariser, leading to better generalisation than competing methods. This analysis underscores the importance of thoroughly investigating the design decisions for auxiliary tasks in diffusion models, such as timestep selection and ensembling. These decisions can have a significant impact on the model’s performance.

Our findings provide important new insights and may inform the development of new approaches leveraging powerful diffusion models for medical imaging tasks. In future work, the performance of TEDM and similar approaches should be compared to the emerging foundation model techniques, where the pre-training is executed at a larger scale than semi-supervised methods. Here, the ability of diffusion models to efficiently capture the data distribution from extensive, unlabelled data holds a promise to overcome the persistent data scarcity problem in medical image segmentation.

\iffalse

\section{Instructions}
\label{sec:instructions}

This is the template for the \textbf{Proceedings Track} for the Machine Learning for Health (ML4H) symposium 2023.
Please follow the below instructions:

\begin{enumerate}
    \item The submission in the Proceedings Paper Track is limited to 8 pages.
    \item Please, use the packages automatically loaded (amsmath, amssymb, natbib, graphicx, url, algorithm2e) to manage references, write equations, and include figures and algorithms. The use of different packages could create problems in the generation of the camera-ready version. Please, follow the example provided in this file.
    \item References must be included in a .bib file.
    \item Please write your paper in a single .tex file.
    \item The manuscript, data and code must be anonymized during the review process.
    \item For writing guidelines please consider the official ML4H call for papers at \url{https://ml4health.github.io/2023/}
\end{enumerate}

\newpage

\section{Introduction}
\label{sec:intro}

This is a sample article that uses the \textsf{jmlr} class with
the \texttt{wcp} class option.  Please follow the guidelines in
this sample document as it can help to reduce complications when
combining the articles into a book. Please avoid using obsolete
commands, such as \verb|\rm|, and obsolete packages, such as
\textsf{epsfig}.\footnote{See
\url{http://www.ctan.org/pkg/l2tabu}} Some packages that are known
to cause problems for the production editing process are checked for
by the \textsf{jmlr} class and will generate an error. (If you want
to know more about the production editing process, have a look at
the video tutorials for the production editors at
\url{http://www.dickimaw-books.com/software/makejmlrbookgui/videos/}.)

Please also ensure that your document will compile with PDF\LaTeX.
If you have an error message that's puzzling you, first check for it
at the UK TUG FAQ
\url{https://texfaq.org/FAQ-man-latex}.  If
that doesn't help, create a minimal working example (see
\url{https://www.dickimaw-books.com/latex/minexample/}) and post
to somewhere like \TeX\ on StackExchange
(\url{http://tex.stackexchange.com/}) or the \LaTeX\ Community Forum
(\url{http://www.latex-community.org/forum/}).

\begin{note}
This is an numbered theorem-like environment that was defined in
this document's preamble.
\end{note}

\subsection{Sub-sections}

Sub-sections are produced using \verb|\subsection|.

\subsubsection{Sub-sub-sections}

Sub-sub-sections are produced using \verb|\subsubsection|.

\paragraph{Sub-sub-sub-sections}

Sub-sub-sub-sections are produced using \verb|\paragraph|.
These are unnumbered with a running head.

\subparagraph{Sub-sub-sub-sub-sections}

Sub-sub-sub-sub-sections are produced using \verb|\subparagraph|.
These are unnumbered with a running head.

\section{Cross-Referencing}

Always use \verb|\label| and \verb|\ref| (or one of the commands
described below) when cross-referencing.  For example, the next
section is Section~\ref{sec:math} but you can also refer to it using
\sectionref{sec:math}. The \textsf{jmlr} class
provides some convenient cross-referencing commands:
\verb|\sectionref|, \verb|\equationref|, \verb|\tableref|,
\verb|\figureref|, \verb|\algorithmref|, \verb|\theoremref|,
\verb|\lemmaref|, \verb|\remarkref|, \verb|\corollaryref|,
\verb|\definitionref|, \verb|\conjectureref|, \verb|\axiomref|,
\verb|\exampleref| and \verb|\appendixref|. The argument of these
commands may either be a single label or a comma-separated list
of labels. Examples:

Referencing sections: \sectionref{sec:math} or
\sectionref{sec:intro,sec:math} or
\sectionref{sec:intro,sec:math,sec:tables,sec:figures}.

Referencing equations: \equationref{eq:trigrule} or
\equationref{eq:trigrule,eq:df} or
\equationref{eq:trigrule,eq:f,eq:df,eq:y}.

Referencing tables: \tableref{tab:operatornames} or
\tableref{tab:operatornames,tab:example} or
\tableref{tab:operatornames,tab:example,tab:example-booktabs}.

Referencing figures: \figureref{fig:nodes} or
\figureref{fig:nodes,fig:teximage} or
\figureref{fig:nodes,fig:teximage,fig:subfigex} or
\figureref{fig:circle,fig:square}.

Referencing algorithms: \algorithmref{alg:gauss} or
\algorithmref{alg:gauss,alg:moore} or
\algorithmref{alg:gauss,alg:moore,alg:net}.

Referencing theorem-like environments: \theoremref{thm:eigenpow},
\lemmaref{lem:sample}, \remarkref{rem:sample}, 
\corollaryref{cor:sample}, \definitionref{def:sample},
\conjectureref{con:sample}, \axiomref{ax:sample} and
\exampleref{ex:sample}.

Referencing appendices: \appendixref{apd:first} or
\appendixref{apd:first,apd:second}.

\section{Equations}
\label{sec:math}

The \textsf{jmlr} class loads the \textsf{amsmath} package, so
you can use any of the commands and environments defined there.
(See the \textsf{amsmath} documentation for further
details.\footnote{Either \texttt{texdoc amsmath} or
\url{http://www.ctan.org/pkg/amsmath}})

Unnumbered single-lined equations should be displayed using
\verb|\[| and \verb|\]|. For example:
\[E = m c^2\]
or you can use the \texttt{displaymath} environment:
\begin{displaymath}
E = m c^2
\end{displaymath}
Numbered single-line equations should be displayed using the
\texttt{equation} environment. For example:
\begin{equation}\label{eq:trigrule}
\cos^2\theta + \sin^2\theta \equiv 1
\end{equation}
This can be referenced using \verb|\label| and \verb|\equationref|.
For example, \equationref{eq:trigrule}.

Multi-lined numbered equations should be displayed using the
\texttt{align} environment.\footnote{For reasons why you 
shouldn't use the obsolete \texttt{eqnarray} environment, see
Lars Madsen, \emph{Avoid eqnarray!} TUGboat 33(1):21--25, 2012.} For example:
\begin{align}
f(x) &= x^2 + x\label{eq:f}\\
f'(x) &= 2x + 1\label{eq:df}
\end{align}
Unnumbered multi-lined equations can be displayed using the
\texttt{align*} environment. For example:
\begin{align*}
f(x) &= (x+1)(x-1)\\
&= x^2 - 1
\end{align*}
If you want to mix numbered with unnumbered lines use the
\texttt{align} environment and suppress unwanted line numbers with
\verb|\nonumber|. For example:
\begin{align}
y &= x^2 + 3x - 2x + 1\nonumber\\
&= x^2 + x + 1\label{eq:y}
\end{align}
An equation that is too long to fit on a single line can be
displayed using the \texttt{split} environment. 
Text can be embedded in an equation using \verb|\text| or
\verb|\intertext| (as used in \theoremref{thm:eigenpow}).
See the \textsf{amsmath} documentation for further details.

\subsection{Operator Names}
\label{sec:op}

Predefined operator names are listed in \tableref{tab:operatornames}.
For additional operators, either use \verb|\operatorname|,
for example $\operatorname{var}(X)$ or declare it with
\verb|\DeclareMathOperator|, for example
\begin{verbatim}
\DeclareMathOperator{\var}{var}
\end{verbatim}
and then use this new command. If you want limits that go above and
below the operator (like \verb|\sum|) use the starred versions
(\verb|\operatorname*| or \verb|\DeclareMathOperator*|).

\begin{table*}[htbp]
\floatconts
  {tab:operatornames}%
  {\caption{Predefined Operator Names (taken from 
   \textsf{amsmath} documentation)}}%
  {%
\begin{tabular}{rlrlrlrl}
\cs{arccos} & $\arccos$ &  \cs{deg} & $\deg$ &  \cs{lg} & $\lg$ &  \cs{projlim} & $\projlim$ \\
\cs{arcsin} & $\arcsin$ &  \cs{det} & $\det$ &  \cs{lim} & $\lim$ &  \cs{sec} & $\sec$ \\
\cs{arctan} & $\arctan$ &  \cs{dim} & $\dim$ &  \cs{liminf} & $\liminf$ &  \cs{sin} & $\sin$ \\
\cs{arg} & $\arg$ &  \cs{exp} & $\exp$ &  \cs{limsup} & $\limsup$ &  \cs{sinh} & $\sinh$ \\
\cs{cos} & $\cos$ &  \cs{gcd} & $\gcd$ &  \cs{ln} & $\ln$ &  \cs{sup} & $\sup$ \\
\cs{cosh} & $\cosh$ &  \cs{hom} & $\hom$ &  \cs{log} & $\log$ &  \cs{tan} & $\tan$ \\
\cs{cot} & $\cot$ &  \cs{inf} & $\inf$ &  \cs{max} & $\max$ &  \cs{tanh} & $\tanh$ \\
\cs{coth} & $\coth$ &  \cs{injlim} & $\injlim$ &  \cs{min} & $\min$ \\
\cs{csc} & $\csc$ &  \cs{ker} & $\ker$ &  \cs{Pr} & $\Pr$
\end{tabular}\par
\begin{tabular}{rlrl}
\cs{varlimsup} & $\varlimsup$ 
& \cs{varinjlim} & $\varinjlim$\\
\cs{varliminf} & $\varliminf$ 
& \cs{varprojlim} & $\varprojlim$
\end{tabular}
}
\end{table*}

\section{Vectors and Sets}
\label{sec:vec}

Vectors should be typeset using \cs{vec}. For example $\vec{x}$.
(The original version of \cs{vec} can also be accessed using
\cs{orgvec}, for example $\orgvec{x}$.)
The \textsf{jmlr} class also provides \cs{set} to typeset a
set. For example $\set{S}$.

\section{Floats}
\label{sec:floats}

Floats, such as figures, tables and algorithms, are moving
objects and are supposed to float to the nearest convenient
location. Please don't force them to go in a particular place. In
general it's best to use the \texttt{htbp} specifier and don't
put the figure or table in the middle of a paragraph (that is
make sure there's a paragraph break above and below the float).
Floats are supposed to have a little extra space above and below
them to make them stand out from the rest of the text. This extra
spacing is put in automatically and shouldn't need modifying.

If your article will later be reprinted in the Challenges for
Machine Learning, please be aware that the CiML books use a
different paper size, so if you want to resize any images use a
scale relative to the line width (\verb|\linewidth|), text width
(\verb|\textwidth|) or text height (\verb|\textheight|).

To ensure consistency, please \emph{don't} try changing the format of the caption by doing
something like:
\begin{verbatim}
\caption{\textit{A Sample Caption.}}
\end{verbatim}
or
\begin{verbatim}
\caption{\em A Sample Caption.}
\end{verbatim}
You can, of course, change the font for individual words or 
phrases, for example:
\begin{verbatim}
\caption{A Sample Caption With Some \emph{Emphasized Words}.}
\end{verbatim}

\subsection{Tables}
\label{sec:tables}

Tables should go in the \texttt{table} environment. Within this
environment use \verb|\floatconts| (defined by \textsf{jmlr})
to set the caption correctly and center the table contents.
The location of the caption depends on the \verb|tablecaption|
setting in the document class options.

\begin{table}[htbp]
 % The first argument is the label.
 % The caption goes in the second argument, and the table contents
 % go in the third argument.
\floatconts
  {tab:example}%
  {\caption{An Example Table}}%
  {\begin{tabular}{ll}
  \bfseries Dataset & \bfseries Result\\
  Data1 & 0.12345\\
  Data2 & 0.67890\\
  Data3 & 0.54321\\
  Data4 & 0.09876
  \end{tabular}}
\end{table}

If you want horizontal rules you can use the \textsf{booktabs}
package which provides the commands \verb|\toprule|, 
\verb|\midrule| and \verb|\bottomrule|. For example, see
\tableref{tab:example-booktabs}.

\begin{table}[hbtp]
\floatconts
  {tab:example-booktabs}
  {\caption{A Table With Horizontal Lines}}
  {\begin{tabular}{ll}
  \toprule
  \bfseries Dataset & \bfseries Result\\
  \midrule
  Data1 & 0.12345\\
  Data2 & 0.67890\\
  Data3 & 0.54321\\
  Data4 & 0.09876\\
  \bottomrule
  \end{tabular}}
\end{table}

If you really want vertical lines as well, you can't use the
\textsf{booktabs} commands as there'll be some unwanted gaps.
Instead you can use \LaTeX's \verb|\hline|, but the rows may
appear a bit cramped.  You can add extra space above or below a
row using \verb|\abovestrut| and \verb|\belowstrut|. For example,
see \tableref{tab:example-hline}. However, you might want to read
the \textsf{booktabs} documentation regarding the use of vertical
lines.

\begin{table}[htbp]
\floatconts
  {tab:example-hline}
  {\caption{A Table With Horizontal and Vertical Lines}}%
  {%
    \begin{tabular}{|l|l|}
    \hline
    \abovestrut{2.2ex}\bfseries Dataset & \bfseries Result\\\hline
    \abovestrut{2.2ex}Data1 & 0.12345\\
    Data2 & 0.67890\\
    Data3 & 0.54321\\
    \belowstrut{0.2ex}Data4 & 0.09876\\\hline
    \end{tabular}
  }
\end{table}

If you want to align numbers on their decimal point, you can
use the \textsf{siunitx} package. For further details see the
\textsf{siunitx} documentation\footnote{Either \texttt{texdoc
siunitx} or \url{http://www.ctan.org/pkg/siunitx}}.

If the table is too wide, you can adjust the inter-column
spacing by changing the value of \verb|\tabcolsep|. For
example:
\begin{verbatim}
\setlength{\tabcolsep}{3pt}
\end{verbatim}
If the table is very wide but not very long, you can use the
\texttt{sidewaystable} environment defined in the
\textsf{rotating} package (so use \verb|\usepackage{rotating}|).
If the table is too long to fit on a page, you can use the
\texttt{longtable} environment defined in the \textsf{longtable}
package (so use \verb|\usepackage{longtable}|).

\subsection{Figures}
\label{sec:figures}

Figures should go in the \texttt{figure} environment. Within this
environment, use \verb|\floatconts| to correctly position the
caption and center the image. Use \verb|\includegraphics|
for external graphics files but omit the file extension. Do not
use \verb|\epsfig| or \verb|\psfig|. If you want to scale the
image, it's better to use a fraction of the line width rather
than an explicit length. For example, see \figureref{fig:nodes}.

\begin{figure}[htbp]
 % Caption and label go in the first argument and the figure contents
 % go in the second argument
\floatconts
  {fig:nodes}
  {\caption{Example Image}}
  {\includegraphics[width=0.5\linewidth]{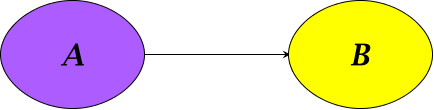}}
\end{figure}

If your image is made up of \LaTeX\ code (for example, commands
provided by the \textsf{pgf} package) you can include it using
\cs{includeteximage} (defined by the \textsf{jmlr} class). This
can be scaled and rotated in the same way as \cs{includegraphics}.
For example, see \figureref{fig:teximage}.

\begin{figure}[htbp]
\floatconts
  {fig:teximage}
  {\caption{Image Created Using \LaTeX\ Code}}
  {\includeteximage[angle=45]{images/teximage}}
\end{figure}

If the figure is too wide to fit on the page, you can use the
\texttt{sidewaysfigure} environment defined in the
\textsf{rotating} package.

Don't use \verb|\graphicspath|.\footnote{This is specific to the
\textsf{jmlr} class, not a general recommendation. The main file
that generates the proceedings or the CiML book is typically in a
different directory to the imported articles, so it modifies the
graphics path when it imports an article.} If the images 
are contained in a subdirectory, specify this when you include the image, for
example \verb|\includegraphics{figures/mypic}|.

\subsubsection{Sub-Figures}
\label{sec:subfigures}

Sub-figures can be created using \verb|\subfigure|, which is
defined by the \textsf{jmlr} class. The optional argument allows
you to provide a subcaption. The label should be placed in the
mandatory argument of \verb|\subfigure|. You can reference the
entire figure, for example \figureref{fig:subfigex}, or you can
reference part of the figure using \verb|\figureref|, for example
\figureref{fig:circle}. Alternatively you can reference the
subfigure using \verb|\subfigref|, for example
\subfigref{fig:circle,fig:square} in \figureref{fig:subfigex}.

\begin{figure}[htbp]
\floatconts
  {fig:subfigex}
  {\caption{An Example With Sub-Figures.}}
  {%
    \subfigure[A Circle]{\label{fig:circle}%
      \includegraphics[width=0.2\linewidth]{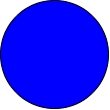}}%
    \qquad
    \subfigure[A Square]{\label{fig:square}%
      \includegraphics[width=0.2\linewidth]{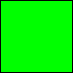}}
  }
\end{figure}

By default, the sub-figures are aligned on the baseline.
This can be changed using the second optional argument
of \verb|\subfigure|. This may be \texttt{t} (top), \texttt{c}
(centered) or \texttt{b} (bottom). For example, the subfigures
\subfigref{fig:circle2,fig:square2} in \figureref{fig:subfigex2}
both have \verb|[c]| as the second optional argument.

\begin{figure}[htbp]
\floatconts
  {fig:subfigex2}
  {\caption{Another Example With Sub-Figures.}}
  {%
    \subfigure[A Small Circle][c]{\label{fig:circle2}%
      \includegraphics[width=0.1\linewidth]{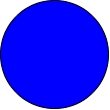}}%
    \qquad
    \subfigure[A Square][c]{\label{fig:square2}%
      \includegraphics[width=0.2\linewidth]{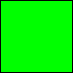}}
  }
\end{figure}

\subsection{Sub-Tables}
\label{sec:subtables}
There is an analogous command \verb|\subtable| for sub-tables.
It has the same syntax as \verb|\subfigure| described above.
You can reference the table using \verb|\tableref|, for example
\tableref{tab:subtabex} or you can reference part of the table,
for example \tableref{tab:ab}. Alternatively you can reference the
subtable using \verb|\subtabref|, for example
\subtabref{tab:ab,tab:cd} in \tableref{tab:subtabex}.

\begin{table}[htbp]
\floatconts
 {tab:subtabex}
 {\caption{An Example With Sub-Tables}}
 {%
   \subtable{%
     \label{tab:ab}%
     \begin{tabular}{cc}
     \bfseries A & \bfseries B\\
     1 & 2
     \end{tabular}
   }\qquad
   \subtable{%
     \label{tab:cd}%
     \begin{tabular}{cc}
     \bfseries C & \bfseries D\\
     3 & 4\\
     5 & 6
     \end{tabular}
   }
 }
\end{table}

By default, the sub-tables are aligned on the top.
This can be changed using the second optional argument
of \verb|\subtable|. This may be \texttt{t} (top), \texttt{c}
(centered) or \texttt{b} (bottom). For example, the sub-tables
\subtabref{tab:ab2,tab:cd2} in \tableref{tab:subtabex2}
both have \verb|[c]| as the second optional argument.

\begin{table}[htbp]
\floatconts
 {tab:subtabex2}
 {\caption{Another Example With Sub-Tables}}
 {%
   \subtable[][c]{%
     \label{tab:ab2}%
     \begin{tabular}{cc}
     \bfseries A & \bfseries B\\
     1 & 2
     \end{tabular}
   }\qquad
   \subtable[][c]{%
     \label{tab:cd2}%
     \begin{tabular}{cc}
     \bfseries C & \bfseries D\\
     3 & 4\\
     5 & 6
     \end{tabular}
   }
 }
\end{table}

\subsection{Algorithms}
\label{sec:algorithms}

Enumerated textual algorithms can be displayed using the
\texttt{algorithm} environment. Within this environment, use
\verb|\caption| to set the caption and you can use an
\texttt{enumerate} or nested \texttt{enumerate} environments.
For example, see \algorithmref{alg:gauss}. Note that algorithms
float like figures and tables.

\begin{algorithm}[htbp]
\floatconts
  {alg:gauss}%
  {\caption{The Gauss-Seidel Algorithm}}%
{%
\begin{enumerate}
  \item For $k=1$ to maximum number of iterations
    \begin{enumerate}
      \item For $i=1$ to $n$
        \begin{enumerate}
        \item $x_i^{(k)} = 
          \frac{b_i - \sum_{j=1}^{i-1}a_{ij}x_j^{(k)}
          - \sum_{j=i+1}^{n}a_{ij}x_j^{(k-1)}}{a_{ii}}$
        \item If $\|\vec{x}^{(k)}-\vec{x}^{(k-1)} < \epsilon\|$,
          where $\epsilon$ is a specified stopping criteria, stop.
      \end{enumerate}
    \end{enumerate}
\end{enumerate}
}%
\end{algorithm}

If you'd rather have the same numbering throughout the algorithm
but still want the convenient indentation of nested 
\texttt{enumerate} environments, you can use the
\texttt{enumerate*} environment provided by the \textsf{jmlr}
class. For example, see \algorithmref{alg:moore}.

\begin{algorithm}
\floatconts
  {alg:moore}%
  {\caption{Moore's Shortest Path}}%
{%
Given a connected graph $G$, where the length of each edge is 1:
\begin{enumerate*}
  \item Set the label of vertex $s$ to 0
  \item Set $i=0$
  \begin{enumerate*}
    \item \label{step:locate}Locate all unlabelled vertices 
          adjacent to a vertex labelled $i$ and label them $i+1$
    \item If vertex $t$ has been labelled,
    \begin{enumerate*}
      \item[] the shortest path can be found by backtracking, and 
      the length is given by the label of $t$.
    \end{enumerate*}
    otherwise
    \begin{enumerate*}
      \item[] increment $i$ and return to step~\ref{step:locate}
    \end{enumerate*}
  \end{enumerate*}
\end{enumerate*}
}%
\end{algorithm}

Pseudo code can be displayed using the \texttt{algorithm2e}
environment. This is defined by the \textsf{algorithm2e} package
(which is automatically loaded) so check the \textsf{algorithm2e}
documentation for further details.\footnote{Either \texttt{texdoc
algorithm2e} or \url{http://www.ctan.org/pkg/algorithm2e}}
For an example, see \algorithmref{alg:net}.

\begin{algorithm2e}
\caption{Computing Net Activation}
\label{alg:net}
 % older versions of algorithm2e have \dontprintsemicolon instead
 % of the following:
 %\DontPrintSemicolon
 % older versions of algorithm2e have \linesnumbered instead of the
 % following:
 %\LinesNumbered
\KwIn{$x_1, \ldots, x_n, w_1, \ldots, w_n$}
\KwOut{$y$, the net activation}
$y\leftarrow 0$\;
\For{$i\leftarrow 1$ \KwTo $n$}{
  $y \leftarrow y + w_i*x_i$\;
}
\end{algorithm2e}

\section{Description Lists}

The \textsf{jmlr} class also provides a description-like 
environment called \texttt{altdescription}. This has an
argument that should be the widest label in the list. Compare:
\begin{description}
\item[add] A method that adds two variables.
\item[differentiate] A method that differentiates a function.
\end{description}
with
\begin{altdescription}{differentiate}
\item[add] A method that adds two variables.
\item[differentiate] A method that differentiates a function.
\end{altdescription}

\section{Theorems, Lemmas etc}
\label{sec:theorems}

The following theorem-like environments are predefined by
the \textsf{jmlr} class: \texttt{theorem}, \texttt{example},
\texttt{lemma}, \texttt{proposition}, \texttt{remark}, 
\texttt{corollary}, \texttt{definition}, \texttt{conjecture}
and \texttt{axiom}. You can use the \texttt{proof} environment
to display the proof if need be, as in \theoremref{thm:eigenpow}.

\begin{theorem}[Eigenvalue Powers]\label{thm:eigenpow}
If $\lambda$ is an eigenvalue of $\vec{B}$ with eigenvector
$\vec{\xi}$, then $\lambda^n$ is an eigenvalue of $\vec{B}^n$
with eigenvector $\vec{\xi}$.
\begin{proof}
Let $\lambda$ be an eigenvalue of $\vec{B}$ with eigenvector
$\xi$, then
\begin{align*}
\vec{B}\vec{\xi} &= \lambda\vec{\xi}
\intertext{premultiply by $\vec{B}$:}
\vec{B}\vec{B}\vec{\xi} &= \vec{B}\lambda\vec{\xi}\\
\Rightarrow \vec{B}^2\vec{\xi} &= \lambda\vec{B}\vec{\xi}\\
&= \lambda\lambda\vec{\xi}\qquad
\text{since }\vec{B}\vec{\xi}=\lambda\vec{\xi}\\
&= \lambda^2\vec{\xi}
\end{align*}
Therefore true for $n=2$. Now assume true for $n=k$:
\begin{align*}
\vec{B}^k\vec{\xi} &= \lambda^k\vec{\xi}
\intertext{premultiply by $\vec{B}$:}
\vec{B}\vec{B}^k\vec{\xi} &= \vec{B}\lambda^k\vec{\xi}\\
\Rightarrow \vec{B}^{k+1}\vec{\xi} &= \lambda^k\vec{B}\vec{\xi}\\
&= \lambda^k\lambda\vec{\xi}\qquad
\text{since }\vec{B}\vec{\xi}=\lambda\vec{\xi}\\
&= \lambda^{k+1}\vec{\xi}
\end{align*}
Therefore true for $n=k+1$. Therefore, by induction, true for all
$n$.
\end{proof}
\end{theorem}

\begin{lemma}[A Sample Lemma]\label{lem:sample}
This is a lemma.
\end{lemma}

\begin{remark}[A Sample Remark]\label{rem:sample}
This is a remark.
\end{remark}

\begin{corollary}[A Sample Corollary]\label{cor:sample}
This is a corollary.
\end{corollary}

\begin{definition}[A Sample Definition]\label{def:sample}
This is a definition.
\end{definition}

\begin{conjecture}[A Sample Conjecture]\label{con:sample}
This is a conjecture.
\end{conjecture}

\begin{axiom}[A Sample Axiom]\label{ax:sample}
This is an axiom.
\end{axiom}

\begin{example}[An Example]\label{ex:sample}
This is an example.
\end{example}

\section{Color vs Grayscale}
\label{sec:color}

It's helpful if authors supply grayscale versions of their
images in the event that the article is to be incorporated into
a black and white printed book. With external PDF, PNG or JPG
graphic files, you just need to supply a grayscale version of the
file. For example, if the file is called \texttt{myimage.png},
then the gray version should be \texttt{myimage-gray.png} or
\texttt{myimage-gray.pdf} or \texttt{myimage-gray.jpg}. You don't
need to modify your code. The \textsf{jmlr} class checks for
the existence of the grayscale version if it is print mode 
(provided you have used \verb|\includegraphics| and haven't
specified the file extension).

You can use \verb|\ifprint| to determine which mode you are in.
For example, in \figureref{fig:nodes}, the 
\ifprint{dark gray}{purple} ellipse represents an input and the
\ifprint{light gray}{yellow} ellipse represents an output.
Another example: {\ifprint{\bfseries}{\color{red}}important text!}

You can use the class option \texttt{gray} to see how the
document will appear in gray scale mode. \textcolor{blue}{Colored
text} will automatically be converted to gray scale in print mode.

The \textsf{jmlr} class loads the \textsf{xcolor}
package, so you can also define your own colors. For example:
\ifprint
  {\definecolor{myred}{gray}{0.5}}%
  {\definecolor{myred}{rgb}{0.5,0,0}}%
\textcolor{myred}{XYZ}.

The \textsf{xcolor} class is loaded with the \texttt{x11names}
option, so you can use any of the x11 predefined colors (listed
in the \textsf{xcolor} documentation\footnote{either 
\texttt{texdoc xcolor} or \url{http://www.ctan.org/pkg/xcolor}}).

\section{Citations and Bibliography}
\label{sec:cite}

The \textsf{jmlr} class automatically loads \textsf{natbib}
and automatically sets the bibliography style, so you don't need to
use \verb|\bibliographystyle|.
This sample file has the citations defined in the accompanying
BibTeX file \texttt{jmlr-sample.bib}. For a parenthetical
citation use \verb|\citep|. For example
\citep{guyon-elisseeff-03}. For a textual citation use
\verb|\citet|. For example \citet{guyon2007causalreport}. 
Both commands may take a comma-separated list, for example
\citet{guyon-elisseeff-03,guyon2007causalreport}.

These commands have optional arguments and have a starred
version. See the \textsf{natbib} documentation for further
details.\footnote{Either \texttt{texdoc natbib} or
\url{http://www.ctan.org/pkg/natbib}}

The bibliography is displayed using \verb|\bibliography|.

\acks{Acknowledgements go here.}

\bibliography{jmlr-sample}

\appendix

\section{First Appendix}\label{apd:first}

This is the first appendix.

\section{Second Appendix}\label{apd:second}

This is the second appendix.

\fi

\FloatBarrier

\acks{This research has been conducted using data from the UK Biobank. M. Rosnati is supported by UK Research and Innovation [UKRI Centre for Doctoral Training in AI for Healthcare grant number EP/S023283/1]. M. Roschewitz  is supported by an Imperial College President’s PhD scholarship.
}

\bibliography{rosnati23}

\begin{thebibliography}{23}
\providecommand{\natexlab}[1]{#1}
\providecommand{\url}[1]{\texttt{#1}}
\expandafter\ifx\csname urlstyle\endcsname\relax
  \providecommand{\doi}[1]{doi: #1}\else
  \providecommand{\doi}{doi: \begingroup \urlstyle{rm}\Url}\fi

\bibitem[Alfaro-Almagro et~al.(2018)Alfaro-Almagro, Jenkinson, Bangerter,
  Andersson, Griffanti, Douaud, Sotiropoulos, Jbabdi, Hernandez-Fernandez,
  Vallee, et~al.]{alfaro2018image}
Fidel Alfaro-Almagro, Mark Jenkinson, Neal~K Bangerter, Jesper~LR Andersson,
  Ludovica Griffanti, Gwena{\"e}lle Douaud, Stamatios~N Sotiropoulos, Saad
  Jbabdi, Moises Hernandez-Fernandez, Emmanuel Vallee, et~al.
\newblock Image processing and quality control for the first 10,000 brain
  imaging datasets from uk biobank.
\newblock \emph{Neuroimage}, 166:\penalty0 400--424, 2018.

\bibitem[Aviles-Rivero et~al.(2022)Aviles-Rivero, Runkel, Papadakis, Kourtzi,
  and Sch{\"o}nlieb]{aviles2022multi}
Angelica~I Aviles-Rivero, Christina Runkel, Nicolas Papadakis, Zoe Kourtzi, and
  Carola-Bibiane Sch{\"o}nlieb.
\newblock Multi-modal hypergraph diffusion network with dual prior for
  alzheimer classification.
\newblock In \emph{Medical Image Computing and Computer Assisted
  Intervention--MICCAI 2022: 25th International Conference, Singapore,
  September 18--22, 2022, Proceedings, Part III}, pages 717--727. Springer,
  2022.

\bibitem[Bakas et~al.(2017)Bakas, Akbari, Sotiras, Bilello, Rozycki, Kirby,
  Freymann, Farahani, and Davatzikos]{bakas2017advancing}
Spyridon Bakas, Hamed Akbari, Aristeidis Sotiras, Michel Bilello, Martin
  Rozycki, Justin~S Kirby, John~B Freymann, Keyvan Farahani, and Christos
  Davatzikos.
\newblock Advancing the cancer genome atlas glioma mri collections with expert
  segmentation labels and radiomic features.
\newblock \emph{Scientific data}, 4\penalty0 (1):\penalty0 1--13, 2017.

\bibitem[Bakas et~al.(2018)Bakas, Reyes, Jakab, Bauer, Rempfler, Crimi,
  Shinohara, Berger, Ha, Rozycki, et~al.]{bakas2018identifying}
Spyridon Bakas, Mauricio Reyes, Andras Jakab, Stefan Bauer, Markus Rempfler,
  Alessandro Crimi, Russell~Takeshi Shinohara, Christoph Berger, Sung~Min Ha,
  Martin Rozycki, et~al.
\newblock Identifying the best machine learning algorithms for brain tumor
  segmentation, progression assessment, and overall survival prediction in the
  brats challenge.
\newblock \emph{arXiv preprint arXiv:1811.02629}, 2018.

\bibitem[Baranchuk et~al.(2021)Baranchuk, Voynov, Rubachev, Khrulkov, and
  Babenko]{baranchuklabel}
Dmitry Baranchuk, Andrey Voynov, Ivan Rubachev, Valentin Khrulkov, and Artem
  Babenko.
\newblock Label-efficient semantic segmentation with diffusion models.
\newblock In \emph{International Conference on Learning Representations}, 2021.

\bibitem[Chaitanya et~al.(2020)Chaitanya, Erdil, Karani, and
  Konukoglu]{chaitanya2020contrastive}
Krishna Chaitanya, Ertunc Erdil, Neerav Karani, and Ender Konukoglu.
\newblock Contrastive learning of global and local features for medical image
  segmentation with limited annotations.
\newblock \emph{Advances in Neural Information Processing Systems},
  33:\penalty0 12546--12558, 2020.

\bibitem[Chen et~al.(2020)Chen, Kornblith, Norouzi, and Hinton]{chen2020simple}
Ting Chen, Simon Kornblith, Mohammad Norouzi, and Geoffrey Hinton.
\newblock A simple framework for contrastive learning of visual
  representations.
\newblock In \emph{International conference on machine learning}, pages
  1597--1607. PMLR, 2020.

\bibitem[Deja et~al.(2023)Deja, Trzcinski, and Tomczak]{deja2023learning}
Kamil Deja, Tomasz Trzcinski, and Jakub~M Tomczak.
\newblock Learning data representations with joint diffusion models.
\newblock \emph{arXiv preprint arXiv:2301.13622}, 2023.

\bibitem[Ho et~al.(2020)Ho, Jain, and Abbeel]{ho2020denoising}
Jonathan Ho, Ajay Jain, and Pieter Abbeel.
\newblock Denoising diffusion probabilistic models.
\newblock \emph{Advances in Neural Information Processing Systems},
  33:\penalty0 6840--6851, 2020.

\bibitem[Jaeger et~al.(2014)Jaeger, Candemir, Antani, W{\'a}ng, Lu, and
  Thoma]{jaeger2014two}
Stefan Jaeger, Sema Candemir, Sameer Antani, Y{\`\i}-Xi{\'a}ng~J W{\'a}ng,
  Pu-Xuan Lu, and George Thoma.
\newblock Two public chest {X}-ray datasets for computer-aided screening of
  pulmonary diseases.
\newblock \emph{Quantitative imaging in medicine and surgery}, 4\penalty0
  (6):\penalty0 475, 2014.

\bibitem[Kim and Ye(2022)]{kim2022diffusion}
Boah Kim and Jong~Chul Ye.
\newblock Diffusion deformable model for 4d temporal medical image generation.
\newblock In \emph{Medical Image Computing and Computer Assisted
  Intervention--MICCAI 2022: 25th International Conference, Singapore,
  September 18--22, 2022, Proceedings, Part I}, pages 539--548. Springer, 2022.

\bibitem[Menze et~al.(2014)Menze, Jakab, Bauer, Kalpathy-Cramer, Farahani,
  Kirby, Burren, Porz, Slotboom, Wiest, et~al.]{menze2014multimodal}
Bjoern~H Menze, Andras Jakab, Stefan Bauer, Jayashree Kalpathy-Cramer, Keyvan
  Farahani, Justin Kirby, Yuliya Burren, Nicole Porz, Johannes Slotboom, Roland
  Wiest, et~al.
\newblock The multimodal brain tumor image segmentation benchmark (brats).
\newblock \emph{IEEE transactions on medical imaging}, 34\penalty0
  (10):\penalty0 1993--2024, 2014.

\bibitem[Patenaude et~al.(2011)Patenaude, Smith, Kennedy, and
  Jenkinson]{patenaude2011bayesian}
Brian Patenaude, Stephen~M Smith, David~N Kennedy, and Mark Jenkinson.
\newblock A bayesian model of shape and appearance for subcortical brain
  segmentation.
\newblock \emph{Neuroimage}, 56\penalty0 (3):\penalty0 907--922, 2011.

\bibitem[Peng et~al.(2022)Peng, Guo, Zhou, Patel, and
  Chellappa]{peng2022towards}
Cheng Peng, Pengfei Guo, S~Kevin Zhou, Vishal~M Patel, and Rama Chellappa.
\newblock Towards performant and reliable undersampled mr reconstruction via
  diffusion model sampling.
\newblock In \emph{Medical Image Computing and Computer Assisted
  Intervention--MICCAI 2022: 25th International Conference, Singapore,
  September 18--22, 2022, Proceedings, Part VI}, pages 623--633. Springer,
  2022.

\bibitem[Pinaya et~al.(2022)Pinaya, Graham, Gray, Da~Costa, Tudosiu, Wright,
  Mah, MacKinnon, Teo, Jager, et~al.]{pinaya2022fast}
Walter~HL Pinaya, Mark~S Graham, Robert Gray, Pedro~F Da~Costa, Petru-Daniel
  Tudosiu, Paul Wright, Yee~H Mah, Andrew~D MacKinnon, James~T Teo, Rolf Jager,
  et~al.
\newblock Fast unsupervised brain anomaly detection and segmentation with
  diffusion models.
\newblock In \emph{Medical Image Computing and Computer Assisted
  Intervention--MICCAI 2022: 25th International Conference, Singapore,
  September 18--22, 2022, Proceedings, Part VIII}, pages 705--714. Springer,
  2022.

\bibitem[Ronneberger et~al.(2015)Ronneberger, Fischer, and
  Brox]{ronneberger2015u}
Olaf Ronneberger, Philipp Fischer, and Thomas Brox.
\newblock U-net: Convolutional networks for biomedical image segmentation.
\newblock In \emph{Medical Image Computing and Computer-Assisted
  Intervention--MICCAI 2015: 18th International Conference, Munich, Germany,
  October 5-9, 2015, Proceedings, Part III 18}, pages 234--241. Springer, 2015.

\bibitem[Rosnati et~al.(2022)Rosnati, Ribeiro, Monteiro, de~Castro, and
  Glocker]{rosnati2022analysing}
Margherita Rosnati, Fabio De~Sousa Ribeiro, Miguel Monteiro, Daniel~Coelho
  de~Castro, and Ben Glocker.
\newblock Analysing the effectiveness of a generative model for semi-supervised
  medical image segmentation.
\newblock In \emph{Machine Learning for Health}, pages 290--310. PMLR, 2022.

\bibitem[Sohl-Dickstein et~al.(2015)Sohl-Dickstein, Weiss, Maheswaranathan, and
  Ganguli]{sohl2015deep}
Jascha Sohl-Dickstein, Eric Weiss, Niru Maheswaranathan, and Surya Ganguli.
\newblock Deep unsupervised learning using nonequilibrium thermodynamics.
\newblock In \emph{International Conference on Machine Learning}, pages
  2256--2265. PMLR, 2015.

\bibitem[Tang et~al.(2019)Tang, Tang, Xiao, and Summers]{tang2019xlsor}
You-Bao Tang, Yu-Xing Tang, Jing Xiao, and Ronald~M Summers.
\newblock Xlsor: A robust and accurate lung segmentor on chest {X}-rays using
  criss-cross attention and customized radiorealistic abnormalities generation.
\newblock In \emph{International Conference on Medical Imaging with Deep
  Learning}, pages 457--467. PMLR, 2019.

\bibitem[Van~Ginneken et~al.(2006)Van~Ginneken, Stegmann, and
  Loog]{van2006segmentation}
Bram Van~Ginneken, Mikkel~B Stegmann, and Marco Loog.
\newblock Segmentation of anatomical structures in chest radiographs using
  supervised methods: a comparative study on a public database.
\newblock \emph{Medical image analysis}, 10\penalty0 (1):\penalty0 19--40,
  2006.

\bibitem[Wang et~al.(2017)Wang, Peng, Lu, Lu, Bagheri, and
  Summers]{wang2017chestx}
Xiaosong Wang, Yifan Peng, Le~Lu, Zhiyong Lu, Mohammadhadi Bagheri, and
  Ronald~M Summers.
\newblock {ChestX-ray8}: Hospital-scale chest {X}-ray database and benchmarks
  on weakly-supervised classification and localization of common thorax
  diseases.
\newblock In \emph{Proceedings of the IEEE conference on computer vision and
  pattern recognition}, pages 2097--2106, 2017.

\bibitem[Wolleb et~al.(2022)Wolleb, Bieder, Sandk{\"u}hler, and
  Cattin]{wolleb2022diffusion}
Julia Wolleb, Florentin Bieder, Robin Sandk{\"u}hler, and Philippe~C Cattin.
\newblock Diffusion models for medical anomaly detection.
\newblock In \emph{Medical Image Computing and Computer Assisted
  Intervention--MICCAI 2022: 25th International Conference, Singapore,
  September 18--22, 2022, Proceedings, Part VIII}, pages 35--45. Springer,
  2022.

\bibitem[Xie and Li(2022)]{xie2022measurement}
Yutong Xie and Quanzheng Li.
\newblock Measurement-conditioned denoising diffusion probabilistic model for
  under-sampled medical image reconstruction.
\newblock In \emph{Medical Image Computing and Computer Assisted
  Intervention--MICCAI 2022: 25th International Conference, Singapore,
  September 18--22, 2022, Proceedings, Part VI}, pages 655--664. Springer,
  2022.

\end{thebibliography}

\appendix

\section{Methods details}\label{apd:first}

\begin{table*}[!t]
\floatconts
{tab:comp}
{\caption{Methods computational cost}}
{
\setlength{\tabcolsep}{6pt}
\begin{tabular}{@{}lcc}
\toprule
            & Theoretical test-time operations &GMAC \\

\midrule

Sup. Baseline       &N&29.2\\
Global CL           &N&29.2\\
Global \& Local CL  &N&29.2\\
LEDM                &$|S_{\text{LEDM}}|\times N + n_{\text{pixels}}\times N_{\text{MLP}}(|S_{\text{LEDM}}|\times n_{\text{latent}},1)$ &  88.0 \\
LEDMe               &$|S_{\text{TEDM}}|\times N +n_{\text{pixels}}\times N_{\text{MLP}}(|S_{\text{TEDM}}|\times n_{\text{latent}},1)$ & 234.6 \\
TEDM (ours)         &$|S_{\text{TEDM}}|\times N + |S_{\text{TEDM}}|\times n_{\text{pixels}}\times N_{\text{MLP}}( n_{\text{latent}},1)$& 234.6 
\end{tabular}}
\end{table*}

\subsection{UK Biobank data preprocessing}
The UK Biobank brains dataset contains \num{42791} patients' scans. We initially separate the data in three sets, a training set with $n_{train}= $ \num{34230}, a validation set with $n_{val}=$ \num{4280} and a test set of $n_{test}=$ \num{4280} patients. After evaluating some methods with $n_{test}=$ \num{4280} and careful consideration of results variance, we reduced the test set to $n_{test}=$ \num{500} without suffering any drops in metrics accuracy.

All scans have voxel size $1mm^3$ and image size $189\times 233 \times 197$, and are paired with the segmentation of 15 subcortical structures’ volumes from FIRST (FMRIB's Integrated Registration and Segmentation Tool~\cite{patenaude2011bayesian}) segmentation, and brain masks. For more details on the scan preprocessing, please refer to~\citet{alfaro2018image}. 

We preprocess the images by clipping the intensities to [0, \num{1500}] to remove large outliers, then normalise the brain pixels using the brain masks so that the $1^{st}$ and $99^{th}$ quantiles correspond to -1 and 1 respectively:
\begin{align}
    x_{norm}[mask\neq0] = a\cdot x[mask\neq0]+b\\
    \text{such that } a = \frac{2}{x^{99\%} - x^{1\%}} \text{ and }b = 1 - a\cdot x^{99\%}
\end{align}
    where $x^{1\%}$ and $x^{99\%}$ are the $1^{st}$ and $99^{th}$ quantiles of $x[mask\neq0]$.

We then split the image and segmentation in $189$ 2D slices, and discard all slices where no brain structures are present in the segmentation, resulting in roughly 100 2D slices per brain image.

\subsection{BraTS data preprocessing}
The BraTS dataset consists of \num{338} patients' scans. For each patient, four scanner modalities are available, ``native T1, post-contrast T1-weighted (T1Gd), T2-weighted (T2), and T2 Fluid Attenuated Inversion Recovery (T2-FLAIR) volumes"\footnote{\url{https://www.med.upenn.edu/cbica/brats2020/data.html}}. Segmentation maps for GD-enhancing tumour, the peritumoural oedema, and the necrotic and non-enhancing tumour core are provided. In addition, the scans are co-registered, resampled to $1mm^3$ resolution as skull stripped. For more information about the BraTS dataset preprocessing, please refer to~\cite{bakas2018identifying,menze2014multimodal}. We separate the data in three sets, a training set with $n_{train}= 269$, a validation set with $n_{val}=36$  and a test set of $n_{test}=33$.  For each scan modality, we calculate the mean and variance of the brain pixels across the training set, excluding the background.
We use the calculated mean and variance to normalise the data distribution to mean 0 and standard deviation 1. 

We then split the images and segmentation in $155$ 2D slices. For each slice, concatenate the four modalities, and take a centre crop of $176\times 176$.

\subsection{Training hyperparameters}
We train the DDPM for \num{100000} steps with batch size 4 and learning rate $\eta = 0.0001$ on a single NVIDIA TITAN X GPU with 12GB capacity. Similarly, we train the Global CL and Global \& Local CL models for \num{100000} steps. All downstream models - the supervised baseline, Global CL and Global \& Local CL fine-tuning, LEDM, LEDMe and TEDM - are trained for \num{20000} steps, with the same learning rate.

\section{Computational cost}
\label{Appendix:comp-cost}
The backbone UNet used across experiments is of $36m$ parameters. We use the package ptflops to estimate the number of operations to $N=29.2M$.
Therefore, Supervised Baseline, Global CL and Global \& Local CL all have a computational cost of $N$.

LEDM requires $|S_{\text{LEDM}}|=|\{50, 150, 250\}|=3$ forward passes through the UNet composing the DDPM backbone of the model, one for each used timestep.
The latent representations extracted from the UNet has $n_{\text{latent}}=960$ dimensions. In the case of LEDM, these dimensions are concatenated and passed through a lightweight multilayer perceptron, composed of three linear layers: input\_channels $\times 128$, $128\times32$ and $32\times$ out\_channels. Here, input\_channels=$|S_{\text{LEDM}}|\times n_{\text{latent}}$ and out\_channels=1. We denote its size by $N_{\text{MLP}}(\text{in\_c, out\_c})$, and note that it is executed $n_{\text{pixels}}$ times. 
LEDMe has a similar complexity structure.

Finally, TEDM, like LEDMe, requires $|S_{\text{tEDM}}|=|\{1, 10, 25, 50, 200, 400, 600, 800\}|=8$ forward passes through the UNet, and requires an MLP of size $N_{\text{MLP}}( n_{\text{latent}},1)$ for each latent representation.
The final numbers for all models can be found in Table~\ref{tab:comp}.

Note that for LEDM, LEDMe and TEDM, the multiple UNet forward passes are the greatest contributors to computational complexity and can be parallelised provided enough computational power, leading to comparable prediction time to the baseline.

\section{Further results and visualisations}\label{apd:second}

\begin{figure*}[!t]
\floatconts
    {fig:temp}
    {\caption{Additional visualisations of segmentations on JSRT, NIH and Montgomery test images as per Figure~\ref{fig:visualsation}. Please zoom in for better visibility of details. }}
    {\includegraphics[width=\textwidth]{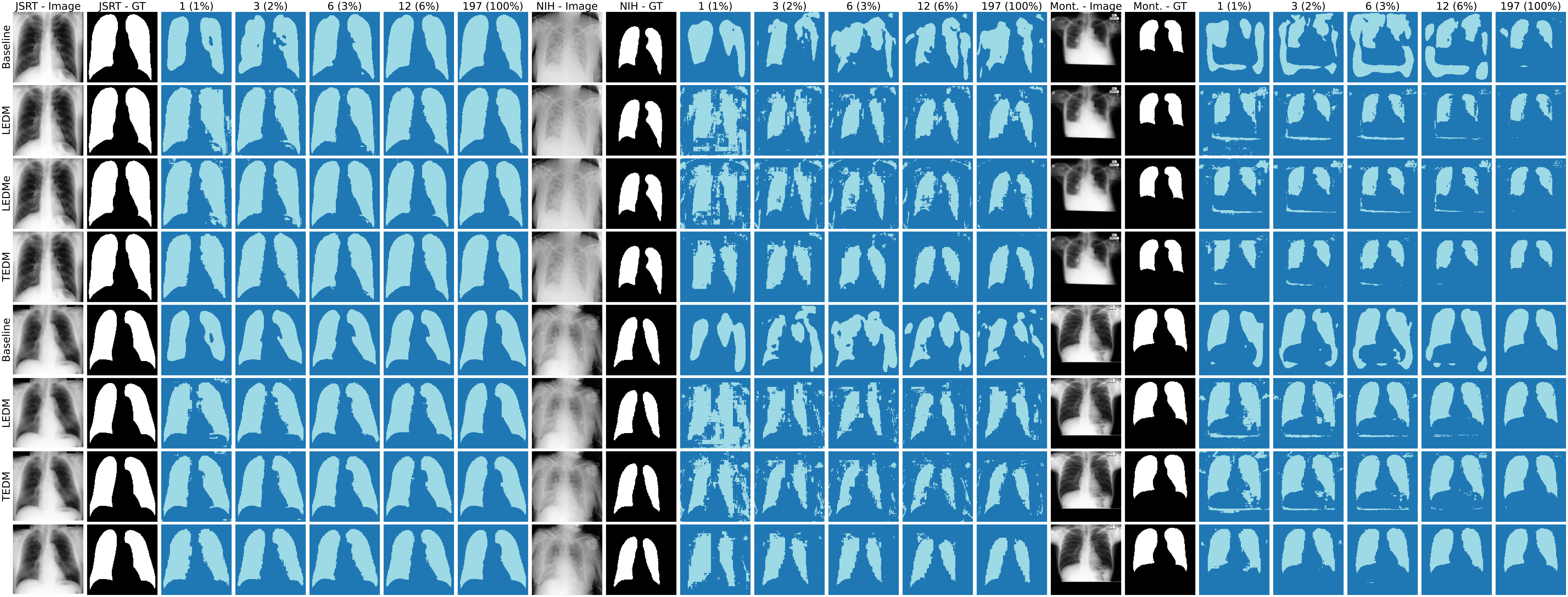}}
\end{figure*}

\begin{figure*}[!t]
\floatconts
  {fig:results_per_timestep_pr}
  {\caption{Additional results on the performance of a logistic regression segmentation
model trained on latent features from individual diffusion steps.
}}
  {\includegraphics[width=.9\textwidth]{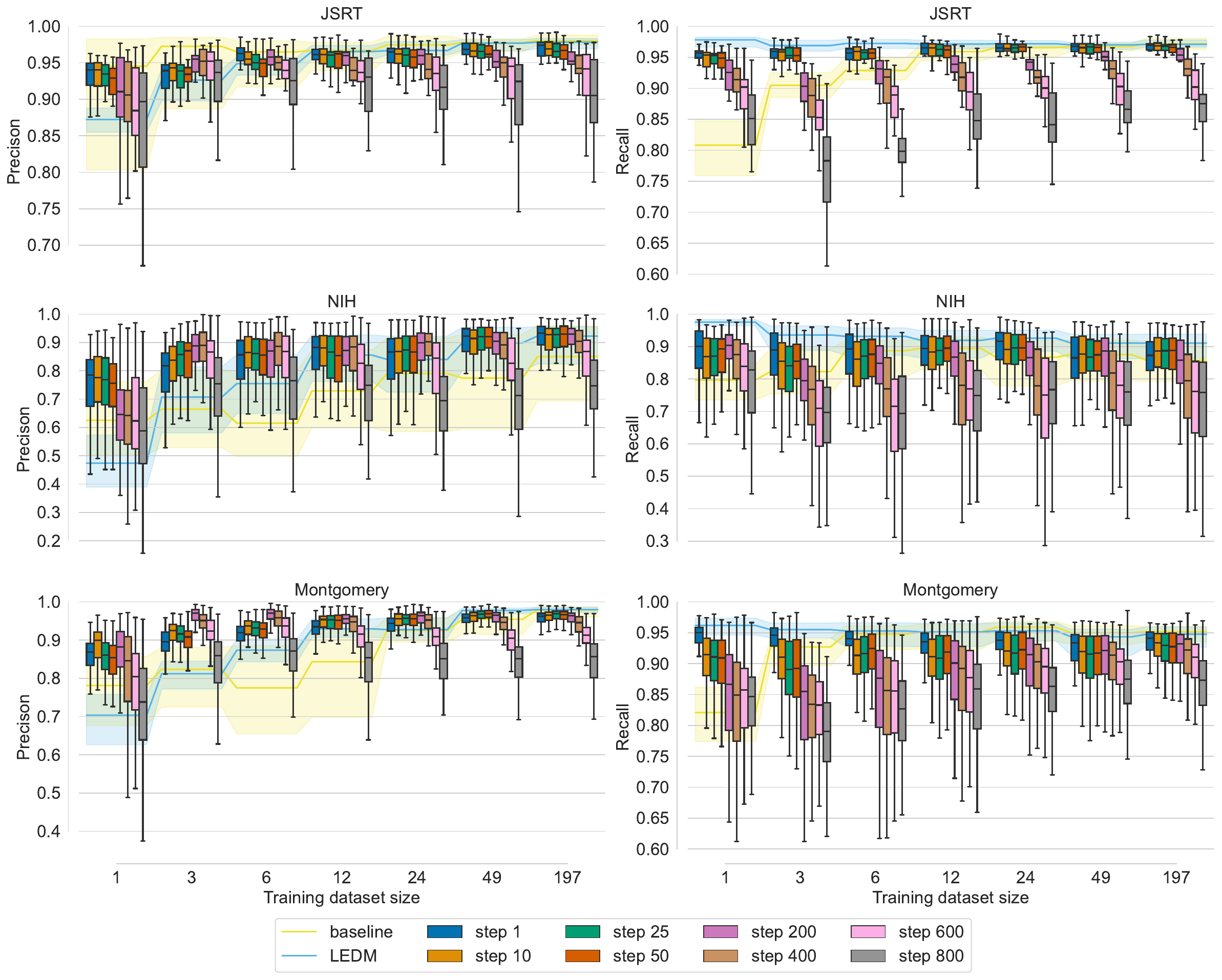}}
  
\end{figure*}

\begin{table*}[!t]
\centering
\caption{Models precision and recall w.r.t.\ ground truth segmentations, as per Table~\ref{tab:res}.}
\label{app:tab:res}
\setlength{\tabcolsep}{2pt}
\begin{tabular}{@{}lccccc@{}}
\toprule
Training size            & 1 & 3 & 6 & 12 & 197 \\

\midrule
&\multicolumn{5}{c}{Precision - JSRT (in-domain for classifier)}                                                                                                                                                                                                                                     \\ 
\cmidrule{2-6}
Sup. Baseline&\textbf{89.2 $\pm$ 12.1}&93.2 $\pm$ 7.2&93.8 $\pm$ 5.9&95.3 $\pm$ 3.7&\textbf{97.9 $\pm$ 1.1}\\
Global CL&86.8 $\pm$ 10.5&95.5 $\pm$ 3.0&\textbf{97.3 $\pm$ 2.6}&\textbf{97.0 $\pm$ 2.0}&\textbf{97.7 $\pm$ 1.5}\\
Global \& Local CL&\textbf{90.2 $\pm$ 9.2}&\textbf{97.1 $\pm$ 2.2}&96.8 $\pm$ 2.0&96.2 $\pm$ 2.0&97.1 $\pm$ 1.6\\
LEDM&85.2 $\pm$ 5.8&91.7 $\pm$ 2.9&94.1 $\pm$ 1.9&96.3 $\pm$ 1.5&97.5 $\pm$ 1.3\\
LEDMe&\textbf{90.1 $\pm$ 4.6}&93.6 $\pm$ 2.3&96.2 $\pm$ 2.0&96.6 $\pm$ 1.6&\textbf{97.9 $\pm$ 0.9}\\
TEDM (ours)&\textbf{91.3 $\pm$ 7.2}&95.4 $\pm$ 2.9&95.6 $\pm$ 2.0&96.4 $\pm$ 1.7&97.5 $\pm$ 1.2\\

\cmidrule{2-6}
&\multicolumn{5}{c}{Recall - JSRT (in-domain for classifier)}     \\ 
\cmidrule{2-6}
Sup. Baseline&81.5 $\pm$ 6.6	&90.6 $\pm$ 3.4	&93.2 $\pm$ 2.6	&95.4 $\pm$ 2.5	&96.8 $\pm$ 2.2	\\
Global CL&91.8 $\pm$ 3.2	&90.2 $\pm$ 3.0	&90.3 $\pm$ 3.8	&93.8 $\pm$ 1.9	&96.6 $\pm$ 2.3	\\
Global \& Local CL&90.0 $\pm$ 3.2	&89.5 $\pm$ 3.3	&89.5 $\pm$ 3.7	&93.4 $\pm$ 2.4	&\textbf{97.2 $\pm$ 1.8}	\\
LEDM&97.4 $\pm$ 1.1	&96.7 $\pm$ 1.2	&\textbf{97.0 $\pm$ 1.6}	&96.6 $\pm$ 2.1	&96.6 $\pm$ 2.1	\\
LEDMe&\textbf{97.7 $\pm$ 1.0}	&\textbf{97.5 $\pm$ 1.4}	&\textbf{97.1 $\pm$ 1.5}	&\textbf{97.4 $\pm$ 1.3}	&\textbf{97.3 $\pm$ 1.9}	\\
TEDM (ours)&95.4 $\pm$ 2.1	&94.3 $\pm$ 1.6	&96.2 $\pm$ 1.5	&96.9 $\pm$ 1.4	&\textbf{97.2 $\pm$ 1.9}	\\

\midrule
&\multicolumn{5}{c}{Precision - NIH (in-domain for DDPM, OOD for classifier)}  \\ 
\cmidrule{2-6}
Sup. Baseline &63.0 $\pm$ 17.0&65.6 $\pm$ 18.3&63.6 $\pm$ 20.3&72.0 $\pm$ 18.3&80.5 $\pm$ 17.4\\
Global CL &60.8 $\pm$ 17.9&78.7 $\pm$ 15.9&76.0 $\pm$ 20.4&83.2 $\pm$ 14.4&89.4 $\pm$ 13.6\\
Global \& Local CL &65.1 $\pm$ 19.1&\textbf{81.7 $\pm$ 15.2}&\textbf{84.5 $\pm$ 15.4}&81.7 $\pm$ 16.8&88.0 $\pm$ 13.9\\
LEDM &48.4 $\pm$ 13.6&69.4 $\pm$ 14.8&74.7 $\pm$ 14.0&83.0 $\pm$ 11.4&88.4 $\pm$ 9.2\\
LEDMe &56.8 $\pm$ 14.1&69.3 $\pm$ 13.7&77.0 $\pm$ 12.9&79.8 $\pm$ 12.0&90.8 $\pm$ 7.8\\
TEDM (ours) &\textbf{70.5 $\pm$ 13.3}&\textbf{82.0 $\pm$ 10.6}&\textbf{86.3 $\pm$ 9.3}&\textbf{90.4 $\pm$ 6.9}&\textbf{95.3 $\pm$ 3.6}\\
\cmidrule{2-6}
&\multicolumn{5}{c}{Recall - NIH (in-domain for DDPM, OOD for classifier)}    \\ 
\cmidrule{2-6}
Sup. Baseline &77.7 $\pm$ 10.3	&80.5 $\pm$ 12.0	&85.4 $\pm$ 10.1	&87.4 $\pm$ 8.0	&84.2 $\pm$ 9.9	\\
Global CL &88.6 $\pm$ 9.7	&83.6 $\pm$ 8.1	&80.1 $\pm$ 13.6	&87.4 $\pm$ 7.7	&85.3 $\pm$ 8.9	\\
Global \& Local CL &80.9 $\pm$ 14.5	&78.6 $\pm$ 11.7	&78.9 $\pm$ 14.0	&84.0 $\pm$ 11.5	&87.6 $\pm$ 8.5	\\
LEDM &\textbf{96.4 $\pm$ 4.2}	&91.8 $\pm$ 5.5	&91.1 $\pm$ 6.2	&90.2 $\pm$ 6.5	&89.9 $\pm$ 5.5	\\
LEDMe &\textbf{96.3 $\pm$ 3.2}	&\textbf{92.5 $\pm$ 6.2}	&\textbf{91.8 $\pm$ 6.7}	&90.9 $\pm$ 7.2	&89.9 $\pm$ 5.7	\\
TEDM (ours) &95.7 $\pm$ 4.0	&\textbf{92.4 $\pm$ 4.2}	&\textbf{92.9 $\pm$ 4.1}	&\textbf{92.7 $\pm$ 4.4}	&\textbf{90.8 $\pm$ 5.0}	\\

\midrule
&\multicolumn{5}{c}{Precision - Montgomery (OOD for DDPM and classifier)}  \\ 
\cmidrule{2-6}
Sup. Baseline &75.1 $\pm$ 16.4&77.6 $\pm$ 16.1&73.5 $\pm$ 18.6&78.1 $\pm$ 19.0&94.9 $\pm$ 8.9\\
Global CL &68.3 $\pm$ 18.3&86.7 $\pm$ 13.7&88.8 $\pm$ 15.8&89.2 $\pm$ 13.8&93.7 $\pm$ 14.1\\
Global \& Local CL &72.2 $\pm$ 20.9&\textbf{90.1 $\pm$ 12.7}&\textbf{92.2 $\pm$ 11.0}&89.2 $\pm$ 14.4&92.9 $\pm$ 14.7\\
LEDM &68.7 $\pm$ 10.5&79.4 $\pm$ 9.7&85.9 $\pm$ 8.8&92.0 $\pm$ 6.8&97.5 $\pm$ 2.7\\
LEDMe &69.7 $\pm$ 9.2&78.8 $\pm$ 9.1&84.8 $\pm$ 8.5&88.5 $\pm$ 7.3&96.4 $\pm$ 3.7\\
TEDM (ours) &\textbf{88.7 $\pm$ 5.3}&\textbf{90.9 $\pm$ 5.9}&\textbf{93.5 $\pm$ 4.9}&\textbf{96.9 $\pm$ 2.4}&\textbf{98.5 $\pm$ 1.0}\\

\cmidrule{2-6}
&\multicolumn{5}{c}{Recall - Montgomery (OOD for DDPM and classifier)}                                                                                                                                                                                                                                    \\ 
\cmidrule{2-6}
Sup. Baseline &80.9 $\pm$ 7.2	&90.9 $\pm$ 5.9	&93.0 $\pm$ 5.6	&93.0 $\pm$ 5.8	&93.6 $\pm$ 4.8	\\
Global CL &88.7 $\pm$ 7.2	&89.9 $\pm$ 4.8	&90.1 $\pm$ 5.5	&92.8 $\pm$ 5.7	&93.0 $\pm$ 6.5	\\
Global \& Local CL &86.1 $\pm$ 10.9	&88.3 $\pm$ 5.5	&88.4 $\pm$ 6.4	&92.2 $\pm$ 5.9	&93.2 $\pm$ 6.0	\\
LEDM &94.9 $\pm$ 4.7	&94.5 $\pm$ 4.2	&93.9 $\pm$ 4.8	&92.9 $\pm$ 8.3	&92.0 $\pm$ 9.4	\\
LEDMe &\textbf{97.0 $\pm$ 3.5}	&\textbf{96.3 $\pm$ 3.7}	&\textbf{95.3 $\pm$ 4.3}	&\textbf{94.3 $\pm$ 5.1}	&\textbf{94.4 $\pm$ 5.1}	\\
TEDM (ours) &92.9 $\pm$ 6.7	&92.4 $\pm$ 6.9	&93.3 $\pm$ 7.1	&92.8 $\pm$ 7.9	&92.6 $\pm$ 9.1	\\

\bottomrule
\end{tabular}
\end{table*}

\begin{table*}[!t]
\floatconts
{app:tab:biobank_brats}
{\caption{Precision and recall scores on the UK Biobank and BraTS datasets, as per Table~\ref{tab:biobank_brats}}.}
{
\setlength{\tabcolsep}{6pt}
\begin{tabular}{@{}lccccc@{}}
&\multicolumn{5}{c}{UK Biobank ($n^{unlabelled}_{train} =$ \num{34000}, $n_{test} = 500$)}\\ 
\toprule
Training size            & 1 & 3 & 6 & 12 & \num{34000}\\
\midrule
&\multicolumn{5}{c}{Precision}\\ 
\cmidrule{2-6}
Sup. Baseline 	& 67.3 $\pm$ 18.9 & 84.5 $\pm$ 11.4 & 84.0 $\pm$ 10.7 & 85.8 $\pm$ 9.5 & 88.7 $\pm$ 9.0 \\
Global CL 	& 59.3 $\pm$ 23.3 & 83.1 $\pm$ 11.5 & 82.9 $\pm$ 11.1 & 85.2 $\pm$ 9.7 & \textbf{89.4 $\pm$ 8.6} \\
Global \& Local CL 	& 52.3 $\pm$ 22.5 & 75.1 $\pm$ 15.0 & 80.3 $\pm$ 11.5 & 81.7 $\pm$ 10.7 & 88.6 $\pm$ 9.2 \\
LEDM 	& 64.9 $\pm$ 21.3 & 83.2 $\pm$ 9.6 & 84.0 $\pm$ 9.6 & 85.5 $\pm$ 9.1 & 86.9 $\pm$ 8.8 \\
LEDMe 	& 51.3 $\pm$ 19.5 & 86.0 $\pm$ 8.9 & 86.4 $\pm$ 9.2 & 85.9 $\pm$ 9.0 & 88.5 $\pm$ 8.9 \\
TEDM 	& \textbf{85.9 $\pm$ 11.7} & \textbf{88.8 $\pm$ 8.3} & \textbf{86.8 $\pm$ 9.1} & \textbf{87.8 $\pm$ 9.0} & 87.7 $\pm$ 9.2 \\
\cmidrule{2-6}
&\multicolumn{5}{c}{Recall}\\ 
\cmidrule{2-6}
Sup. Baseline 	& 41.3 $\pm$ 20.5 & 67.8 $\pm$ 16.4 & 79.7 $\pm$ 11.4 & \textbf{82.5 $\pm$ 11.2} & \textbf{88.6 $\pm$ 6.4} \\
Global CL 	& 30.6 $\pm$ 19.6 & 70.2 $\pm$ 14.9 & 78.8 $\pm$ 11.3 & \textbf{82.8 $\pm$ 10.4} & 85.8 $\pm$ 9.5 \\
Global \& Local CL 	& 39.6 $\pm$ 19.4 & 73.6 $\pm$ 11.0 & \textbf{81.1 $\pm$ 9.9} & 82.7 $\pm$ 10.1 & 86.6 $\pm$ 7.6 \\
LEDM 	& 64.4 $\pm$ 17.7 & \textbf{76.2 $\pm$ 13.2} & \textbf{81.4 $\pm$ 10.2} & 81.5 $\pm$ 11.2 & 86.2 $\pm$ 8.0 \\
LEDMe 	& \textbf{66.0 $\pm$ 18.1} & 75.2 $\pm$ 12.7 & 79.4 $\pm$ 11.1 & 82.5 $\pm$ 10.7 & 85.0 $\pm$ 8.0 \\
TEDM 	& 58.6 $\pm$ 20.3 & 73.2 $\pm$ 13.3 & 79.7 $\pm$ 11.1 & 80.0 $\pm$ 11.9 & 83.0 $\pm$ 8.6 \\
\bottomrule
&\multicolumn{5}{c}{ } \\ 
&\multicolumn{5}{c}{BraTS ($n^{unlabelled}_{train} = 268$, $n_{test} = 33$)} \\ 
\toprule
Training size            & 1 & 3 & 6 & 12 & \num{33}\\
\midrule%\cmidrule{2-6}
&\multicolumn{5}{c}{Precision}\\ 
\cmidrule{2-6}
Sup. Baseline 	& 25.7 $\pm$ 30.0 & 45.1 $\pm$ 37.4 & 54.6 $\pm$ 37.6 & 62.2 $\pm$ 35.1 & \textbf{74.1 $\pm$ 26.8} \\
Global CL 	& 12.0 $\pm$ 25.3 & 38.6 $\pm$ 34.9 & 48.3 $\pm$ 37.1 & 57.1 $\pm$ 34.9 & 66.6 $\pm$ 29.9 \\
Global \& Local CL 	& 31.6 $\pm$ 35.7 & 40.5 $\pm$ 37.2 & 49.5 $\pm$ 36.3 & 60.7 $\pm$ 35.1 & 66.3 $\pm$ 29.2 \\
LEDM 	& 26.4 $\pm$ 28.5 & 44.5 $\pm$ 37.9 & 56.7 $\pm$ 35.8 & 61.6 $\pm$ 35.0 & 70.6 $\pm$ 27.4 \\
LEDMe 	& 27.9 $\pm$ 29.4 & 51.2 $\pm$ 37.6 & 60.8 $\pm$ 35.2 & 61.4 $\pm$ 34.8 & 70.4 $\pm$ 27.5 \\
TEDM 	& \textbf{46.2 $\pm$ 34.2} & \textbf{61.4 $\pm$ 35.8} & \textbf{67.2 $\pm$ 33.6} & \textbf{67.4 $\pm$ 33.4} & 72.4 $\pm$ 27.0 \\
\cmidrule{2-6}
&\multicolumn{5}{c}{Recall}\\ 
\cmidrule{2-6}
Sup. Baseline 	& 18.9 $\pm$ 28.4 & \textbf{43.7 $\pm$ 36.4} & \textbf{48.1 $\pm$ 35.8} & \textbf{49.5 $\pm$ 35.5} & \textbf{71.1 $\pm$ 26.2} \\
Global CL 	& 13.6 $\pm$ 29.1 & 38.9 $\pm$ 36.2 & 33.1 $\pm$ 33.6 & 45.2 $\pm$ 33.5 & 56.9 $\pm$ 30.9 \\
Global \& Local CL 	& 21.0 $\pm$ 31.3 & 38.3 $\pm$ 35.8 & 40.6 $\pm$ 34.9 & 39.4 $\pm$ 33.3 & 56.8 $\pm$ 31.8 \\
LEDM 	& \textbf{35.8 $\pm$ 26.7} & 37.0 $\pm$ 34.3 & \textbf{45.8 $\pm$ 33.6} & 51.0 $\pm$ 32.0 & 63.8 $\pm$ 26.9 \\
LEDMe 	& 26.8 $\pm$ 26.8 & 36.0 $\pm$ 32.9 & \textbf{46.7 $\pm$ 34.6} & \textbf{53.1 $\pm$ 32.5} & 64.7 $\pm$ 27.7 \\
TEDM 	& 27.6 $\pm$ 28.2 & 37.3 $\pm$ 33.3 & 42.4 $\pm$ 33.3 & 47.9 $\pm$ 32.9 & 59.3 $\pm$ 30.2 \\
\bottomrule

\end{tabular}}
\end{table*}

\end{document}